\documentclass[10pt,twocolumn,letterpaper]{article}

\usepackage{cvpr}              % To produce the CAMERA-READY version
\usepackage{multirow}
\usepackage{pifont}
\usepackage{graphicx}  % 引入graphicx包以使用\scalebox
\usepackage{listings}
\usepackage{subcaption}   % 控制子表格
\usepackage{cuted}
\usepackage{float}
\usepackage{caption}    % For better caption control
\usepackage{soul}
\usepackage{xcolor}
\usepackage{pifont}
\usepackage{tikz}
\usepackage[utf8]{inputenc}
\usepackage{tcolorbox}
\usepackage{xcolor}
\usepackage{pifont}
\usepackage{geometry}
\usepackage{amsmath} % 用于数学符号
\usepackage{amssymb} % 用于符号如 \dag
\usepackage{hyperref}

\usepackage[frozencache,cachedir=minted-cache]{minted}
\usepackage{mdframed} % 用于创建有边框的代码块
\setminted{
  fontsize=\scriptsize,  % 代码字体大小
  baselinestretch=1.1,  % 行距
  breaklines=true,  % 自动换行
  framesep=2mm,  % 代码与边框的距离（可以留作内边距）
  tabsize=2,  % Tab键的空格大小
  bgcolor=gray!10,  % 设置背景颜色为浅灰色
}
\definecolor{customgreen}{rgb}{0.35, 0.65, 0.35}
\definecolor{customred}{rgb}{0.7, 0.35, 0.35}
\definecolor{customblue}{rgb}{0.3, 0.6, 0.8}
\definecolor{customblue1}{rgb}{0.3, 0.6, 0.8}
\definecolor{customblack}{rgb}{0.2, 0.2, 0.2}

\newcommand{\blackcircle}[1]{%
    \begin{tikzpicture}[baseline=-0.9ex]
        \node[circle,fill=black,text=white,inner sep=0.2pt,minimum size=1pt] {#1};
    \end{tikzpicture}%
}

\newcommand{\cmark}{\textcolor{green}{\ding{51}}} % 定义绿色对勾
\newcommand{\xmark}{\textcolor{red}{\ding{55}}}   % 定义红色叉号

\definecolor{darkred}{rgb}{0.6, 0, 0}
\definecolor{darkgreen}{rgb}{0, 0.5, 0}
\definecolor{darkblue}{rgb}{0, 0, 0.55}
\definecolor{cvprblue}{rgb}{0.21,0.49,0.74}
\def\paperID{3012} % *** Enter the Paper ID here
\def\confName{CVPR}
\def\confYear{2025}

\title{
Video-Bench: Human-Aligned Video Generation Benchmark
}

\renewcommand{\fnsymbol}[1]{%
  \ifcase#1\relax
    † % 第一个符号是 †
  \or
    * % 第二个符号是 *
  \else
    \fnsymbol{#1} % 其他符号使用默认顺序
  \fi
}

\author{
    Hui Han$^{1}$\thanks{Equal Contribution.}\footnotemark[1] \hspace{0.6cm} 
    Siyuan Li$^{1}$\footnotemark[1] \hspace{0.6cm}
    Jiaqi Chen$^{2,3,4}$\footnotemark[1] \hspace{0.6cm} 
    Yiwen Yuan$^{5}$\footnotemark[1] \hspace{0.6cm} 
    Yuling Wu$^5$ \\
    Yufan Deng$^6$ \hspace{0.6cm}
    Chak Tou Leong$^7$ \hspace{0.6cm} 
    Hanwen Du$^8$ \hspace{0.6cm} 
    Junchen Fu$^9$ \hspace{0.6cm} 
    Youhua Li$^{10}$ \hspace{1cm} \\
    Jie Zhang$^4$ \hspace{1cm} 
    Chi Zhang$^{11}$ \hspace{1cm} 
    Li-jia Li$^{12}$ \hspace{1cm} 
    Yongxin Ni$^{13}$\thanks{Corresponding author (email: niyongxin@u.nus.edu).}
        \vspace{.5em}
    \\
    $^1$Shanghai Jiao Tong University 
    \quad
    $^2$Stanford University
    \quad
    $^3$Fellou AI
    \quad
    $^4$Fudan University
    \quad \\
    $^5$Carnegie Mellon University 
    \quad
    $^6$Peking University 
    \quad
    $^7$Hong Kong Polytechnic University
    \quad \\
    $^8$Soochow University
    \quad
    $^9$University of Glasgow
    \quad 
    $^{10}$City University of Hong Kong
    \quad \\
    $^{11}$Westlake University
    \quad
    $^{12}$LiveX AI
    \quad
    $^{13}$National University of Singapore
      \vspace{.5em} 
  \\
  \textcolor{black}{\url{https://github.com/Video-Bench/Video-Bench.git}}
}

\begin{document}

\maketitle

% 手动添加符号说明
% \vspace{1em} % 增加垂直间距，使说明与作者列表分开
% \noindent
% \textsuperscript{\dag} Equal contribution. \\
% \textsuperscript{*} Corresponding author.

% \vspace{-50mm}
% \begin{strip}
%     \centering
% \includegraphics[width=1.0\linewidth]{figure/dimension-old.png}
%     % \vspace{-1em}
%     \captionof{figure}{
%         \textbf{Overview of \texttt{Video-Bench}.}
%         \textit{Left}: We introduce comprehensive evaluation dimensions in two main categories: \textcolor{darkred}{\textit{\textbf{video-condition alignment}}} (Sec.~\ref{sec:video_condition_alignment}) and \textcolor{darkgreen}{\textit{\textbf{video quality}}} (Sec.~\ref{sec:video_quality}). 
%         % 
%         \textit{Right}: For these two types of dimensions, we analyze their challenges (Sec.~\ref{sec:challenges}) and corresponding propose
%         % . Then, we propose 
%         the MLLM-based automated video evaluation suite containing two major techniques: \textcolor{darkred}{\textit{\textbf{chain-of-query}}} (Sec.~\ref{sec:chain_of_query}) and \textcolor{darkgreen}{\textit{\textbf{few-shot scoring}}} (Sec.~\ref{sec:few_shot_scoring}). 
%         % We also propose \textcolor{darkblue}{\textit{\textbf{grid-layout}}} for dimensions related to \textcolor{darkblue}{\textit{\textbf{visual dynamics}}} (\textit{i.e.}, Action consistency and temporal consistency).
%     }
%     \label{fig:intro}
% \end{strip}

\begin{strip}
    \centering
    % \vspace*{-20mm}  % 去除上方的空白间距，试着调整这个数值
    \includegraphics[width=\textwidth, height=\textheight, keepaspectratio]{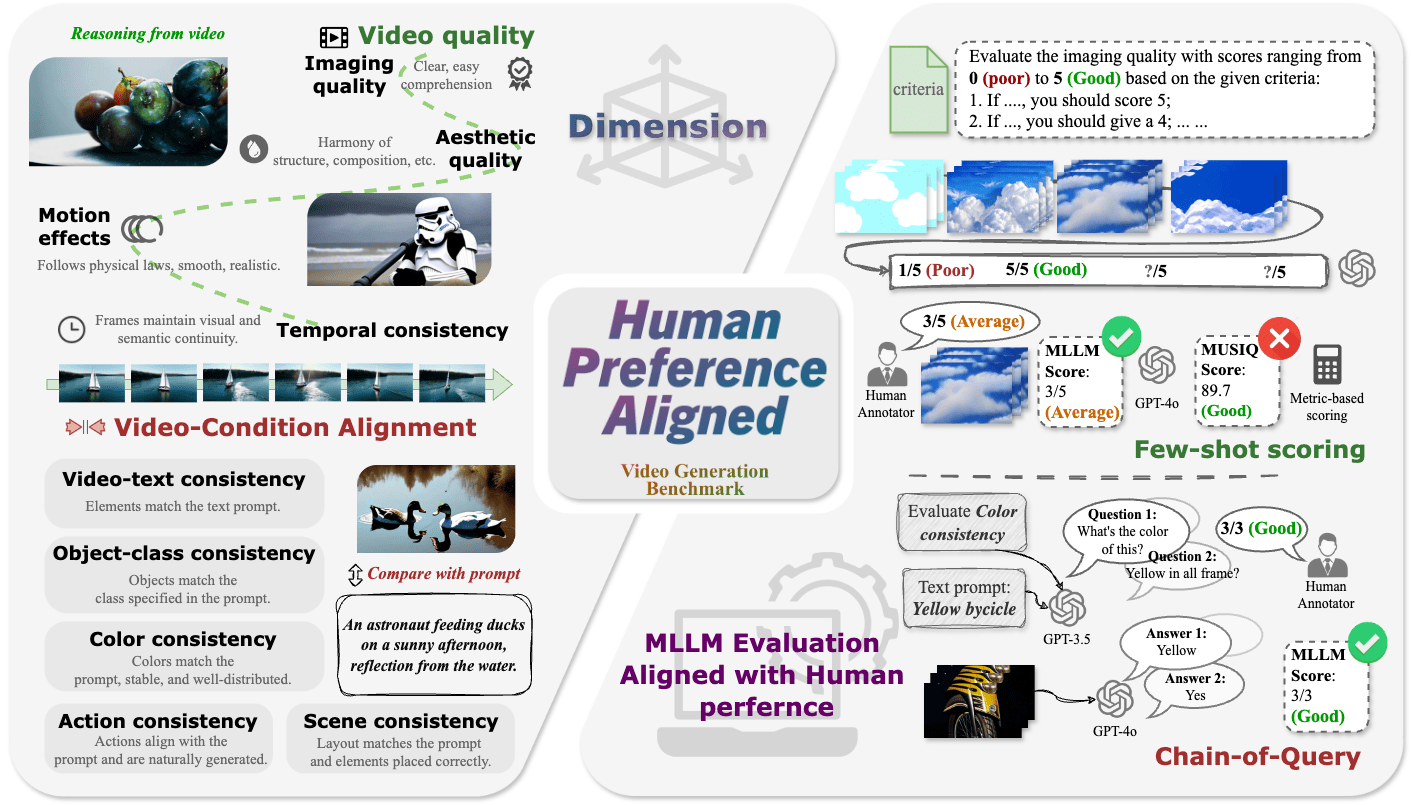}
    \captionof{figure}{
        \textbf{Overview of \texttt{Video-Bench}.}
        \textit{Left}: We introduce comprehensive evaluation dimensions in two main categories: \textcolor{darkred}{\textit{\textbf{video-condition alignment}}} (Sec.~\ref{sec:video_condition_alignment}) and \textcolor{darkgreen}{\textit{\textbf{video quality}}} (Sec.~\ref{sec:video_quality}). 
        \textit{Right}: For these two types of dimensions, we analyze their challenges (Sec.~\ref{sec:challenges}) and corresponding propose the MLLM-based automated video evaluation suite.
    }
    \label{fig:intro}
\end{strip}
% \clearpage

\begin{abstract}
Video generation assessment is essential for ensuring that generative models produce visually realistic, high-quality videos while aligning with human expectations. 
Current video generation benchmarks fall into two main categories: traditional benchmarks, which use metrics and embeddings to evaluate generated video quality across multiple dimensions but often lack alignment with human judgments; and large language model (LLM)-based benchmarks,
though capable of human-like reasoning, are constrained by a limited understanding of video quality metrics and cross-modal consistency.
To address these challenges and establish a benchmark that better aligns with human preferences, this paper introduces \texttt{\textbf{Video-Bench}}, a comprehensive benchmark featuring a rich prompt suite and extensive evaluation dimensions. This benchmark represents the first attempt to systematically leverage MLLMs across all dimensions relevant to video generation assessment in generative models. By incorporating few-shot scoring and chain-of-query techniques, Video-Bench provides a structured, scalable approach to generated video evaluation.
Experiments on advanced models including Sora demonstrate that Video-bench achieve superior alignment with human preferences across all dimensions. Moreover, in instances where our framework’s assessments diverge from human evaluations, it consistently offers more objective and accurate insights, suggesting an even greater potential advantage over traditional human judgment.
\end{abstract}
    
\section{Introduction}
In recent years, generative models have achieved remarkable advancements in the field of visual content generation~\cite{gu2022vector, huang2023collaborative,huang2023reversion,nichol2021glide,podell2023sdxl,rombach2022high,saharia2022photorealistic}.
% \textcolor{red}{[need citations]}.
% 
These breakthroughs have further propelled progress in video generation~\cite{blattmann2023align, he2022latent, hong2022cogvideo, luo2023videofusion, singer2022make, wang2023modelscope, wang2023lavie,wu2023tune, zhang2023show}, enabling unprecedented capabilities in creating dynamic and realistic videos from textual descriptions. 
As video generation models like Sora\footnote{https://openai.com/index/sora/}, Pika\footnote{https://pika.art/}, and Runway\footnote{https://runwayml.com/ai-tools/gen-2-text-to-video/}, continue to evolve, there is an increasingly urgent need for reliable evaluation benchmarks to assess their quality. 
A reliable benchmark should be based on a comprehensive understanding of human preferences for video content: humans tend to prefer videos that are more aligned with input conditions such as color and scene descriptions (\textit{i.e.}, high video-condition alignment), and those that demonstrate better aesthetic quality and temporal consistency (\textit{i.e.}, high video quality).

To capture these aspects of human preferences, existing automated video generation benchmarks can be broadly categorized into two types: 
(1) \textit{\textbf{Metrics and Embedding-based Benchmarks}}: These  benchmarks~\cite{huang2024vbench,liu2023evalcrafter,liu2024fetv} attempt to combine various metrics to improve evaluation effectiveness. Metrics like  Inception Score (IS)~\cite{salimans2016improved}, Fréchet Inception Distance (FID)~\cite{heusel2017gans}, Fréchet Video Distance (FVD)~\cite{unterthiner2018towards, unterthiner2019fvd} are used to calculate video quality~\cite{huang2024vbench}. Additionally, embeddings from pre-trained models such as CLIP~\citep{radford2021learning} or BLIP~\citep{li2023blip} are utilized to measure video-text alignment. 
(2) \textit{\textbf{Large Language Model (LLM)-based Benchmark}}:
LLMs has demonstrated promising results in evaluating text-to-image generation due to their strong language understanding and reasoning abilities~\citep{ku2312viescore, lu2024llmscore, cho2024visual, hu2023tifa, yarom2024you, cho2023davidsonian}, leading a few video-generation benchmarks to integrate LLMs into their evaluation framework~\citep{sun2024t2v, wu2024towards}.

However, both categories face different challenges in aligning with human preferences comprehensively.
Metrics and embedding-based benchmarks, while providing quantitative measurements, often yield evaluation results that significantly misalign with human preferences~\cite{ding2022cogview2,otani2023toward}. In contrast, LLM-based benchmarks, with their strong reasoning abilities, show promising potential in better understanding and simulating human evaluation logic, thus more closely imitating human evaluation patterns.
Nevertheless, current LLM-based approaches encounter two critical limitations: 
In \blackcircle{1} \textcolor{darkred}{\textit{\textbf{video-condition alignment}}} evaluation, challenges arise in performing cross-modal comparisons between textual prompts and video content, 
In \blackcircle{2} \textcolor{darkgreen}{\textit{\textbf{video quality}}} evaluation, difficulties emerge in the ambiguity in converting textual critiques into concrete evaluation scores.  

Overcoming these limitations is crucial for developing reliable evaluation benchmarks that authentically reflect human video preferences.

To address these challenges, this paper presents \texttt{\textbf{Video-Bench}}, a video generation benchmark shown in Figure~\ref{fig:intro} with comprehensive dimension suite and automatic multimodal LLM (MLLM) evaluation policy highly aligned with human preference. 
Building on this foundation, we propose a systematic MLLM-based framework to enhance the model's capability to directly comprehend video content and evaluate its quality. To ensure both depth and clarity in assessment, the protocol incorporates two key techniques: 
\blackcircle{1} \textcolor{darkred}{\textbf{\textit{chain-of-query}}}, which facilitates iterative questioning to conduct a progressively multi-perspective evaluation along the preference dimension; and 
\blackcircle{2} \textcolor{darkgreen}{\textbf{\textit{few-shot scoring}}}, which uses multimodal few-shot examples to guide the direction and scale of the evaluation process. 
Our experimental results show that Video-Bench evaluates video generation models without human intervention. Our proposed Chain-of-query and Few-shot scoring allows us to outperform all past evaluation methods by showing the highest correlation coefficient with human preferences.
The following contributions are made in this paper.
\begin{itemize}
    \item We present a comprehensive benchmark with rich dimensions, a large amount of prompt data, and precise manual annotations. 
    \item We reformulate the video generation evaluation problem from the perspective of a MLLM-based framework, offering an alternative to traditional human evaluation and computational metrics, achieving a high level of alignment with human preferences.
    \item Our experimental results indicate that the proposed framework outperforms existing state-of-the-art evaluation methods.
\end{itemize}

\section{Related work}
\label{rel_work}

\subsection{Visual Evaluation with MLLM}
In Liu \textit{et al.}~\citep{liu2024survey}'s summary, most LLM-based evaluation focus on measuring the alignment with respect to human instructions. GPT-4v Eval~\citep{zhang2023gpt} uses LLM to do single-score grading and pair-wise comparison across four tasks. It shows that with prompting LLM's judgement has high agreement with human evaluation. LLMScore~\citep{lu2024llmscore} transfers images into image-level and object-level descriptions and feeds the descriptions, along with original prompt into LLM to reason about the consistency. VIEScore~\citep{ku2312viescore} feeds rating instructions, considering semantic consistency and percentual quality aspects, prompt and image into MLLM and outputs the scores under each of the aspect. Many LLM-based evaluation focuses on question generation and answering (QG/A)~\citep{hu2023tifa, yarom2024you,cho2023davidsonian}. 
TIFA~\citep{hu2023tifa} first creates binary questions from text prompt using LLM and then measures alignment by calculating the average number of correct answers produced by VQA system. VQ2~\citep{yarom2024you} also leverage VQA system but instead checks for whether the textual answer is accurate. 
\subsection{Video Generation Benchmark}
Traditionally, T2V benchmarks follow either metric-based and model-based benchmarks for evaluation~\citep{liu2024fetv,liu2023evalcrafter, huang2024vbench, kou2024subjective}. For example,
VBench~\citep{huang2024vbench} and EvalCrafter~\citep{liu2023evalcrafter} both propose to evaluate generated videos by scoring them in multiple dimensions under the following categories--quality of video, motion consistency, temporal consistency and text-video alignment. Then the human ablation study is performed on those benchmarks. EvalCrafter~\citep{liu2023evalcrafter} correlates user score with computed metrics by fitting a linear regression model, while VBench~\citep{huang2024vbench} calculates the Pearson and Spearman correlatoin between the two. The recent advancement of Large Language Models (LLM) and Multimodal Large Language Models (MLLM) has bring a promising direction to this field. More recent benchmarks, such as CompBench~\citep{sun2024t2v} and T2VScore~\citep{wu2024towards} take a hybrid approach, only using MLLM to measure alignment with instructions, still using metrics-based methods to evaluate other aspects regarding perception.
\section{Benchmark}

\subsection{Evaluation Dimension Suite}
\label{sec:dimension}
As shown in Figure~\ref{fig:intro}, we divide ``\textit{video generation quality}" into two distinct aspects: ``\textit{video quality}" and ``\textit{video-condition alignment}". The former focuses solely on the quality of the video itself, while the latter assesses whether the video is generated according to the requirements specified by human instructions.
\subsubsection{Video-condition alignment}
\label{sec:video_condition_alignment}
The video-condition alignment is crucial for evaluating content quality, particularly in meeting specific user needs. Different consistency dimensions are scored based on difficulty, using either a three-point scale (1-3) or a five-point scale (1-5)\footnote{For more complex dimensions, the five-point scale is applied to encompass a broader range of criteria.}.
\begin{enumerate}[label=\scalebox{1.2}{\ding{\numexpr171+\arabic*}}]
    \item \textbf{Object Class Consistency} This metric evaluates whether the objects presented in the video match those described in the text prompt. The focus is on whether the objects are generated correctly, are clearly identifiable, and whether their appearance and structure align with objective reality and human perception. In addition, the movement of objects should be examined for abnormal deformations. Scoring is based on a three-point scale.
    \item \textbf{Action Consistency} This metric assesses whether the actions in the video accurately reflect the descriptions in the text prompt. The focus is on the accuracy of generated actions, their clarity, and whether the appearance and process of the actions conform to objective reality and human cognition. Scoring is based on a three-point scale.
    \item \textbf{Color Consistency} This metric measures whether the colors of objects in the video match those described in the text prompt. The colors in the video should remain consistent without sudden changes or discrepancies. Scoring is based on a three-point scale.
    \item \textbf{Scene Consistency} This metric evaluates the alignment of the generated scene with the text prompt, ensuring that all relevant elements are clearly visible and arranged logically, with appearance and structure consistent with reality and human perception. Scoring is based on a three-point scale.
    \item \textbf{Video-text Consistency} This metric assesses the overall consistency between the video and the text prompt, ensuring that all core elements (\textit{e.g.}, humans, animals, actions, objects, scenes, style, spatial relationships, quantity relationships, \textit{etc.}) are accurately represented and that the video quality does not impair user comprehension. Scoring is based on a five-point scale.
\end{enumerate}
\subsubsection{Video Quality}
\label{sec:video_quality}
Video quality is a critical dimension for evaluating the visual fidelity of generated content, encompassing multiple aspects, each focusing on different elements of video quality. 
All dimensions are on a five-point scale (1 - 5). 

\begin{enumerate}[label=\scalebox{1.2}{\ding{\numexpr171+\arabic*}}]
\item \textit{\textbf{Imaging Quality}} This metric focuses on the technical quality of individual frames, assessing visual distortions such as noise, blur, overexposure, or other artifacts that may negatively impact viewer perception. The goal is to ensure that frames are technically as flawless as possible, minimizing defects that could affect overall quality.

\item \textit{\textbf{Aesthetic Quality}} Beyond technical considerations, the aesthetic quality of generated frames is evaluated to determine if they meet aesthetic standards and align with human perceptual expectations. This involves assessing artistic appeal, composition, and overall visual coherence to ensure that the generated content is both visually pleasing and naturally harmonious.

\item \textit{\textbf{Temporal Consistency}} Temporal consistency is a crucial aspect of video quality that directly impacts the smoothness and naturalness of the video: \textit{1) Visual feature consistency}: This metric ensures that visual features such as color, brightness, and texture transition smoothly across consecutive frames, avoiding abrupt changes or inconsistencies, thus maintaining continuity and high perceived quality. \textit{2) Semantic consistency}: This aspect focuses on the consistency of objects, subjects, and scenes across frames, ensuring that objects maintain stable positions, shapes, and appearances, thereby avoiding sudden transformations or unnatural transitions that could detract from the viewing experience.

\item \textit{\textbf{Motion Quality}} Motion quality is fundamental in distinguishing video from static images and plays a vital role in evaluating the dynamics of generated content: \textit{1) Motion rationality}: This metric assesses whether the movement of objects within the video adheres to physical laws and appears realistic. Unrealistic or erratic movements can undermine the sense of immersion and significantly reduce perceived quality.\textit{ 2) Motion amplitude}: This metric assesses whether the extent of movement is appropriate and aligns with the intended actions described in the prompt. Movements should neither be exaggerated nor overly subtle, ensuring that the generated dynamic content appears purposeful and lifelike.
\end{enumerate}
These evaluation criteria rely on text-based guidelines to assess the quality of a video. However, the ambiguity in these textual guidelines and the lack of clear measurement scales make precise evaluation challenging.
\subsection{Prompt Suite}
\label{sec:prompt_suite}
Following the VBench~\cite{huang2024vbench}, we designed a prompt suite for each evaluation dimension, with approximately $70$-$90$ video generation prompts per dimension, with a total of $419$ prompts. 
For 
``Action Consistency", 
``Temporal Consistency", 
and ``Motion Quality", 
we combined human action data from the Kinetics-400 dataset~\cite{kay2017kinetics} with rigid body and animal motion data from Subject consistency subset of VBench~\cite{huang2024vbench}, to achieve a comprehensive and targeted evaluation of dynamic-related dimensions. 
For all other dimensions, 
our prompt suite aligns with the corresponding VBench~\cite{huang2024vbench} prompts. 
To mitigate bias from sampling randomness in video generation models, each prompt was sampled three times in our experiments.
\label{sec:human_annotation}

\section{Evaluation Framework with MLLMs}
\subsection{Challenges}
\label{sec:challenges}
Straight-forward prompting a multimodal language model (MLLM) to rate a video can present significant challenges due to the distinct complexities of video-condition alignment (Sec.~\ref{sec:video_condition_alignment}) and video quality evaluation (Sec.~\ref{sec:video_quality}) assessment. Here, we outline the two primary issues:
\begin{itemize}
    \item \blackcircle{1} 
    \setulcolor{darkred}\ul{\textit{\textbf{Video-condition alignment}: Cross-modal comparisons are difficult for MLLM.}}
    When evaluating the consistency between generated videos and textual prompts, multimodal large language models (MLLMs) are required to compare visual signals with textual concepts. However, this cross-modal comparison is challenging~\cite{peng2024dreambench++}, as MLLMs are prone to textual bias, often generating "hallucinations" and failing to accurately detect inconsistencies between text and video.
    MLLMs are faced with the task of comparing objects of different modalities. MLLMs are prone to hallucination, making incorrect statements about the user-provided image and the prompts~\citep{yin2023woodpecker, huang2024opera, zhou2023analyzing}.
    \item \blackcircle{2} \setulcolor{darkgreen}\ul{\textit{\textbf{Video quality}: Textual critiques are ambiguous.}} For dimensions such as video quality, the challenge lies in requiring an MLLM to assign an absolute score to a video based on human textual critiques. However, the criteria for such evaluations are inherently ambiguous. For example, both MLLMs and humans struggle to precisely distinguish between levels such as ``very blurry", ``somewhat blurry," and ``very clear." 
\end{itemize}
\begin{figure}[t]
    \centering
    \includegraphics[width=1.0\linewidth]{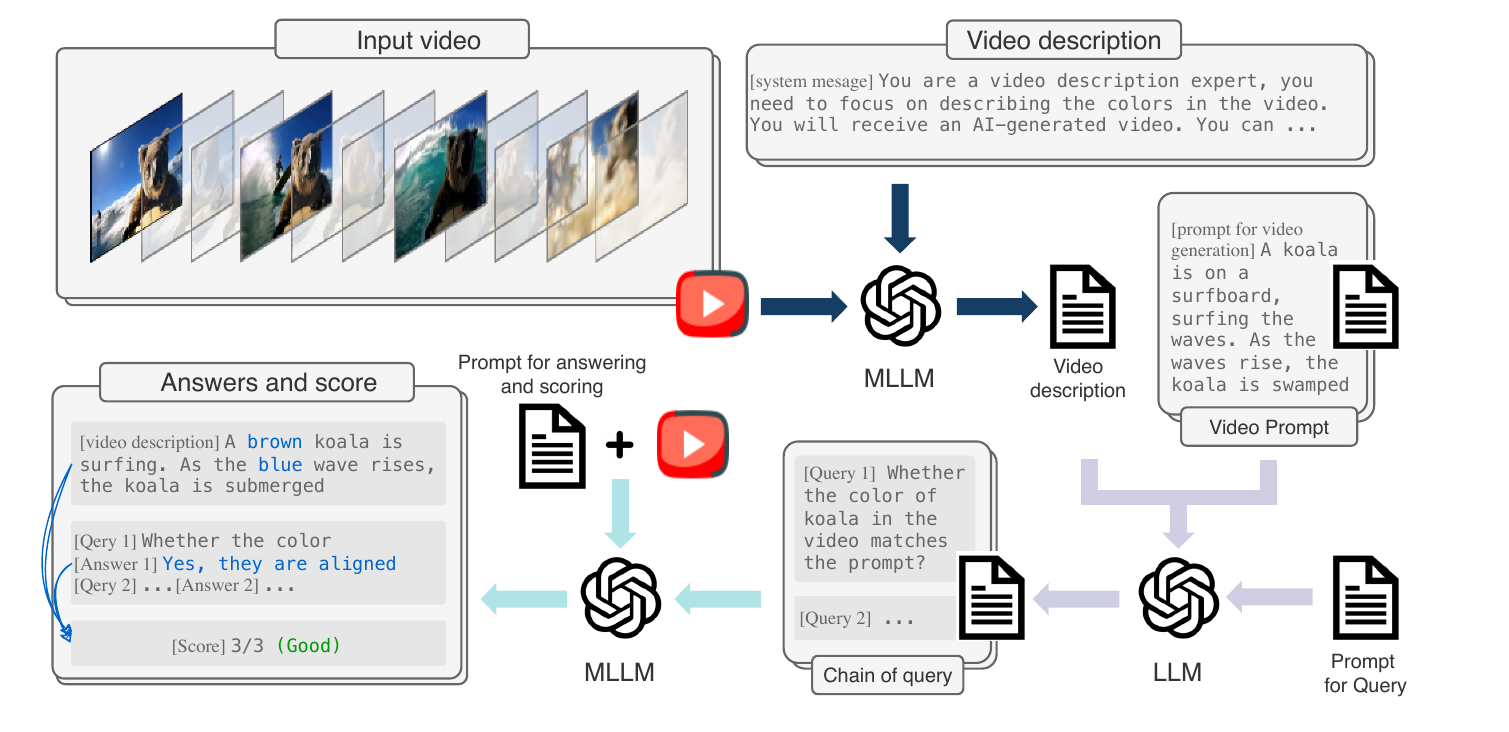}
    \caption{\textbf{\textit{Chain-of-query} for video-condition alignment evaluation.} 
    An iterative process where MLLM transforms video content into text descriptions, enabling detailed assessment of video-condition alignment through multi-turn queries.}
    \label{fig:chain_of_query}
\end{figure}
In our work, we propose \blackcircle{1} \textcolor{darkred}{\textbf{\textit{chain-of-query}}} and \blackcircle{2} \textcolor{darkgreen}{\textbf{\textit{few-shot scoring}}} to address above challenges. 
\subsection{Chain-of-query}
\label{sec:chain_of_query}
For the \setulcolor{darkred}\ul{\textit{\textbf{video-condition alignment}}} (Sec.~\ref{sec:video_condition_alignment}) assessment, traditional LLM-based evaluation~\cite{sun2024t2v} relies on single-turn, direct question-answer methods, which often struggle with cross-modal alignment between video content and textual descriptions. 
This approach can miss critical details and fail to capture nuanced alignment, leading to incomplete evaluations.

To overcome this, we avoid direct cross-modal comparison by first transforming all relevant information from the video modality into a textual form. 
This allows for a more consistent, cross-modality comparison to assess alignment accurately.
The evaluation process begins by generating an initial video description with MLLM based on the video prompt. 
Next, an LLM creates \textbf{\textit{a chain of queries}}, focusing on specific dimensions that need evaluation. 
These queries probe inconsistencies and overlooked aspects in the initial description. 
MLLM then revisits the video, addressing each query to refine the description with richer, dimension-relevant details. 
This iterative multi-turn process, as illustrated in Figure~\ref{fig:chain_of_query}, results in a more comprehensive textual representation, enabling precise scoring for alignment.
Specifically, our chain-of-query includes the following steps:
\begin{enumerate}[label=\scalebox{1.2}{\ding{\numexpr171+\arabic*}}]
    \item \textbf{Video description:}  MLLM is first prompted to generate a full description and a one-sentence summary of the video.
    \item \textbf{Query chain generation:} LLM generates $N$ sets of questions with video description from first step and video generation prompt. The question generation strategy is predefined based on evaluation dimension. For example, for the color dimension. As shown in Figure~\ref{fig:chain_of_query}, the first generated query is  ``\textit{whether the color of koala in the video matches the prompt?}", while the second set focus on the color dominance with respect to other colors in the video: ``\textit{Has the koala's brown color ever been confused with the color of the waves?}".
    \item \textbf{Answer chain generation:} MLLM is then prompted to answer the questions from above step. To ensure MLLM first analyze the whole video before answering individual questions, MLLM is first asked to perform reflection by re-generating a description. Finally MLLM responded “Yes, they are aligned” and ``No, there is no confusion.''
    \item \textbf{Final scoring:} MLLM utilizes both the video content and the multi-turn conversation history, along with textual guidelines, to arrive at a final score. 
\end{enumerate}

\subsection{Few-shot scoring}
\label{sec:few_shot_scoring}
\begin{figure}[t]
    \centering
    \includegraphics[width=1.0\linewidth]{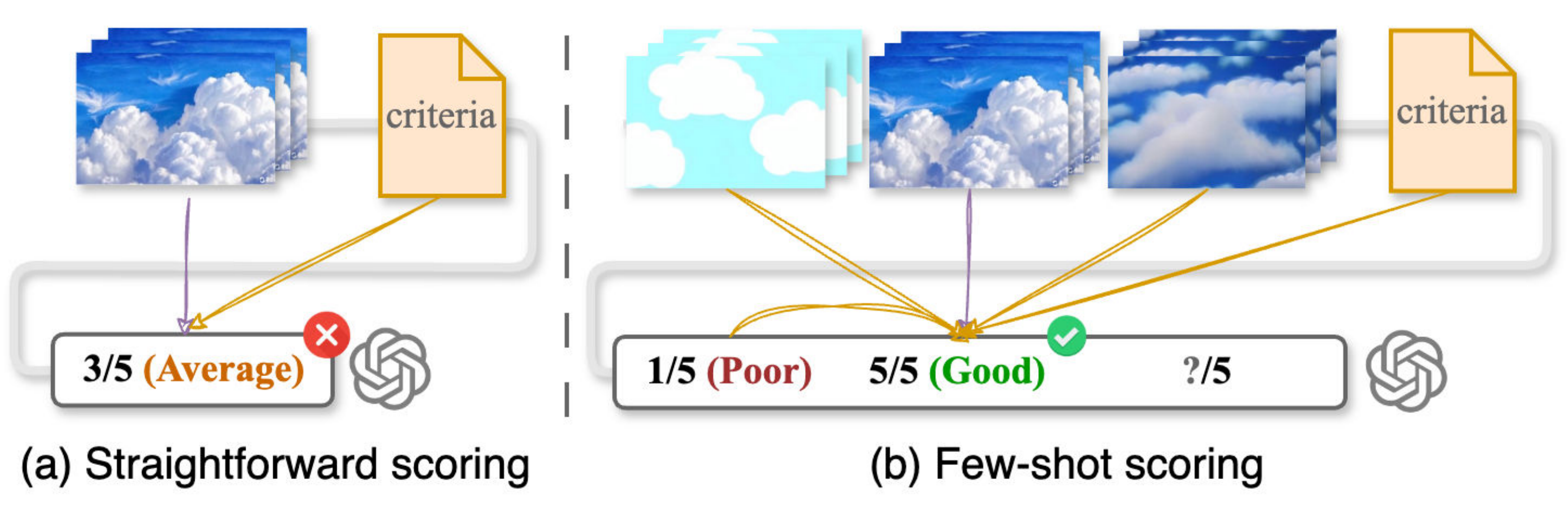}
    \caption{\textbf{Schematic scoring strategy comparison.} (a) Straightforward scoring assigns a single score based on the criteria, resulting in an average rating, while (b) Few-shot scoring leverages multiple examples for calibration, providing a nuanced assessment with scores ranging from poor to good.}
    \label{fig:few_shot_scoring}
\end{figure}
For \setulcolor{darkgreen}\ul{\textbf{\textit{video quality}}} dimensions (Sec.~\ref{sec:video_quality}) such as ``Aesthetic Quality", our scoring criteria include descriptive text for each level, such as ``3 points: moderate aesthetic quality" and ``4 points: good aesthetic quality". 
Although each level has more detailed descriptions, the boundaries between them remain ambiguous. 
In our empirical observations, when we prompt an MLLM to evaluate specific dimensions like video quality, as shown in Figure~\ref{fig:few_shot_scoring} (a), the model often assigns the same (typically average) score to all videos. 
This outcome suggests that the MLLM is not sufficiently sensitive to variations in video quality, likely due to the inherent vagueness in textual criteria alone.
We believe that textual guidelines alone are insufficient and that additional multimodal references are necessary to establish clearer benchmarks. To address this, we batch multiple videos generated from the same prompt, using each video as a contextual reference for the others within the same batch. Specifically, as illustrated in Figure~\ref{fig:few_shot_scoring} (b), when scoring the second video, all videos in the batch, as well as the score of the first video, serve as implicit references, providing a comparative framework that enhances the accuracy of quality assessment across the batch.
\begin{table*}[htbp]
% \vspace{-0.6cm}
\caption{
\textbf{\textbf{\textit{Video-Bench}} Leaderboard.} Higher scores indicate better performance. The best score in each dimension is highlighted in bold. ``Avg Rank" is the average rank of multiple dimensions, the lower the better.
}

% \scriptsize
\renewcommand\tabcolsep{1pt}
\renewcommand\arraystretch{1.2}
\renewcommand{\footnote}{\fnsymbol{footnote}} 
% \small
\setlength{\abovecaptionskip}{0.0cm}
\setlength{\belowcaptionskip}{-0.45cm}
\centering
\hspace*{0.00000000000000001cm}
% \scalebox{0.86}{%
\resizebox{\textwidth}{!}{
\begin{tabular}{l|ccccc|cccccc|c}

\hline

\hline

\hline

\hline

\multirow{3}*{Model} & \multicolumn{5}{c|}{\underline{\textit{\textbf{Video quality}}}} & \multicolumn{6}{c|}{\underline{\textit{\textbf{Video-Condition Alignment}}}} & Overall\\

 & Imaging  & Aesthetic & Temporal & Motion & Avg & Video-text & Object-class & Color & Action & Scene & Avg& Avg\\
& Quality & Quality & Consist. & Effects & Rank& Consist. & Consist. & Consist. & Consist. & Consist. & Rank & Rank\\
\hline

\hline
Gen3~\citep{runwaygen3} &$\textbf{4.66}$ & $\textbf{4.44}$ & $\textbf{4.74}$&$ \textbf{3.99}$& $\textbf{1}$&$4.38$&$2.81$&$2.87$&$2.59$&$\textbf{2.93}$&$ 2$&$\textbf{1}$\\

Cogvideox~\citep{yang2024cogvideox} &$3.87$ &$3.84$  &$4.14$  & $3.55$ & $3$ &$\textbf{4.62}$&$2.81$ &$\textbf{2.92}$ &$\textbf{2.81}$&$\textbf{2.93}$& $\textbf{1}$ &  $2$\\

VideoCrafter2~\citep{chen2024videocrafter2} &$4.08$& $3.85$  &  $3.69$ & $2.81$ & $4$ & $4.18$ &$\textbf{2.85}$& $2.90$ & $2.53$ & $2.78$& $3$ & $3$\\

Kling~\citep{klingkuaishou} & $4.26$ & $3.82$ & $4.38$ & $3.11$& $2$&$4.07$&$2.70$&$2.81$&$2.50$&$2.82$& $5$&$4$ \\

Show-1~\citep{zhang2023show} &$3.30$ &$3.28$ &  $3.90$ &$2.90$& $5$&$4.21$&$2.82$&$2.79$&$2.53$&$2.72$&$4$&$5$\\

LaVie~\citep{wang2023lavie} &$3.00$&$2.94$&$3.00$&$2.43$&$7$&$3.71$&$2.82$&$2.81$&$2.45$&$2.63$& $6$& $6$\\

PiKa-Beta~\citep{pikalab} &$3.78$ &$3.76$&$3.40$& $ 2.59$&$6$&$3.78$&$2.51$&$2.52$&$2.25$&$2.60$&$ 7 $&$ 7$ \\

\hline

\hline

\hline

\hline

\end{tabular}
}
% \vspace{0.25cm}

\label{tab:model_compare}
% \vspace{-0.2cm}
\end{table*}

\begin{table*}[htbp]
\caption{
\textbf{The human alignment score.} This score is measured by Spearman's rank correlation coefficient. Higher score indicates better performance. The best score in each dimension is highlighted in bold. In practice, ComBench$^{*}$~\cite{sun2024t2v} is a reproduction version on our benchmark metrics.
}
%ComBench评估文本一致性指标是在让VLM描述视频帧后，再让其依据视频帧和描述根据评分标准进行评分
\scriptsize
\renewcommand\tabcolsep{2.6pt}
\renewcommand\arraystretch{1.2}
\renewcommand{\footnote}{\fnsymbol{footnote}} 
\small
\setlength{\abovecaptionskip}{0.0cm}
\setlength{\belowcaptionskip}{-0.45cm}
\centering
\hspace*{0.00000000000000001cm}
% \scalebox{0.86}{%
\resizebox{\textwidth}{!}{
\begin{tabular}{l|cccc|ccccc}

\hline

\hline

\hline

\hline

% \multirow{2}*{Method} & \multirow{2}*{Modality} & \multicolumn{2}{c}{Backbones}\vline & \multicolumn{2}{c}{\textit{validation}}\vline & \multicolumn{2}{c}{\textit{test}} \\ 
% & & Image & LiDAR & mAP$\uparrow$ & NDS$\uparrow$ & mAP$\uparrow$ & NDS$\uparrow$ \\
% \hline

% \multirow{2}*{Models} & Imaging & Aesthetic & Temporal & Motion & Overall \\
%  & Quality & Quality & Consistency & Effects & Consistency \\\multicolumn{2}{c}{Models}

% \multirow{2}*{Models} & Imaging & Aesthetic & Temporal & Motion & Overall & Object & \multirow{2}*{Scene} & \multirow{2}*{Color} & \multirow{2}*{Action} & \multirow{2}*{Avg.} \\
% & Quality & Quality & Consistency & Effects & Consistency & Class & & &  \\

\multirow{3}*{Metrics} & \multicolumn{4}{c|}{\underline{\textit{\textbf{Video quality}}}} & \multicolumn{5}{c}{\underline{\textit{\textbf{Video-Condition Alignment}}}} \\

& Imaging  & Aesthetic & Temporal & Motion & Video-text & Object-class & Color & Action & Scene  \\
& Quality & Quality & Consist. & Effects & Consist. & Consist. & Consist. & Consist. & Consist. \\
\hline

\hline

MUSIQ~\citep{ke2021musiq} & $0.363$ & - & - & - & - & - & - & - & - \\

% DOVER~\citep{wu2023exploring}&EvalCrafter~\citep{liu2023evalcrafter} & $0.320$ & $0.445$ & - & - & - & - & - & - & - \\

LAION~\citep{AestheticPredictor} & - & $0.446$ & - & - & - & - & - & - & - \\

CLIP~\citep{radford2021learning} & - & - & $0.260$ & - & - & - & - & - & - \\

% Warping Error~\citep{teed2020raft}& & - & - & $0.216$ & - & - & - & - & - & - \\

RAFT~\citep{teed2020raft} & - & - & - & $0.329$ & - & - & - & - & - \\

Amt~\citep{li2023amt} & - & - & - & $0.329$ & - & - & - & - & - \\

ViCLIP~\citep{wang2023internvid} & - & - & - & - & $0.445$ & - & - & - & - \\

% BLIPScore~\citep{li2022blip}&FETV~\citep{liu2024fetv} & - & - & - & - & $0.290$  & - & - & -& - \\

UMT~\citep{li2023unmasked} & - & - & - & - & - & - & - & $0.411$ & - \\

GRiT~\citep{wu2022grit} & - & - & - & - & - & $0.469$ & $0.545$ & - & - \\

Tag2Text~\citep{huang2023tag2text} & - & - & - & - & - & - & - & - & $0.422$ \\
CompBench~\cite{sun2024t2v}$^{*}$ & - & - & - & - & $0.633$ & $0.611$ & $0.696$ &$ 0.633$ &$ 0.631$ \\
% SAM-Track\citep{cheng2023segment} & - & - & - & - & - & 5 & 5 & - & - \\

\hline

%\hline
%\multicolumn{2}{l|}{Human}  & & & & & & & & & &  \\
%\hline

\hline
\multicolumn{1}{l|}{Ours} & $\textbf{0.733}$ & $\textbf{0.702}$ & $\textbf{0.402}$ & $\textbf{0.514}$ & $\textbf{0.732}$ & $\textbf{0.735}$ & $\textbf{0.750}$ & $\textbf{0.718}$ & $\textbf{0.733}$ \\

\hline

\hline

\hline

\hline

\end{tabular}
}
% \vspace{0.25cm}
\label{tab:human_align}
\end{table*}

\begin{table*}[htbp]
\caption{
\textbf{Inter rater agreement degree (Krippendorff’s $\alpha$).} Higher score indicates better performance. ``HU" stands for human, ``HA" stands for \textbf{\textit{Video-Bench}} and ``GPT" stands for evaluations from single-GPT without chain-of-query, few-shot prompting or grid-view components. 
}

\scriptsize
\renewcommand\tabcolsep{2.1pt}
\renewcommand\arraystretch{1.2}
\renewcommand{\footnote}{\fnsymbol{footnote}} 
\small
\setlength{\abovecaptionskip}{0.0cm}
\setlength{\belowcaptionskip}{-0.45cm}
\centering
\hspace*{0.00000000000000001cm}
% \scalebox{0.86}{%
\begin{tabular}{c|cccc|ccccc|c}

\hline

\hline

\hline

\hline

& \multicolumn{4}{c|}{\underline{\textit{\textbf{Video quality}}}} & \multicolumn{5}{c|}{\underline{\textit{\textbf{Video-Condition Alignment}}}} & \multirow{3}*{\textbf{Avg.}} \\

Entities&Imaging  & Aesthetic & Temporal & Motion & Video-text & Object-class & Color & Action & Scene &  \\
& Quality & Quality & Consistency & Effects & Consistency & Consistency & Consistency & Consistency & Consistency & \\
\hline

\hline
 HU - HU & $0.63$ & $0.55$ &$0.57$ &$0.57$&$0.47$ & $0.51$&$0.55$ & $0.37$ & $0.42$ & $0.52$\\
 HU - GPT & $0.51$ & $0.42$ & $0.45$& $0.35$&$0.47$ &
 $0.50$ & $0.49$ & $ 0.37$ & $0.11$ & $0.41$\\
 HU - HA & $0.61$ & $0.54$ & $0.48$& $0.48$&$0.50$ & $0.52$&$0.54$ & $0.40$ & $0.43$ &$ 0.50$\\
  
\hline

\hline

\hline

\hline

\end{tabular}
% }
 \vspace{0.25cm}

\label{tab:agreement}
\end{table*}

\begin{table*}[htbp]
    \caption{
    \textbf{Ablation study of component design.} Metric: Spearman's rank correlation coefficient. ``Consist." denotes ``Consistency".}
    \centering
\scriptsize

\renewcommand\arraystretch{1.2}
\small
\centering
\hspace*{0.00000000000000001cm}
    \begin{subtable}[t]{0.4\textwidth}  % 左侧子表格
    \renewcommand\tabcolsep{1pt}
        \centering
        \begin{tabular}{c|ccccc}
            \hline

            \hline

            \hline

            \hline
            
            &  \multicolumn{4}{c}{\underline{\textit{\textbf{Video quality}}}} & \multirow{3}*{\textbf{Avg.}}\\

            Few Shot  &Imaging& Aesthetic & Temporal & Motion \\
            Scoring &  Quality & Quality & Consist. & Effects\\
            \hline
            
            \hline
             &  $0.639$ & $0.627$ & $0.526$&$0.452$ & $0.561$ \\
             \checkmark &  $\textbf{0.733}$ &$\textbf{0.702}$&$\textbf{0.531}$ &$\textbf{0.514}$ & $\textbf{0.620}$\\
             \hline

            \hline
                
            \hline
                
            \hline
        \end{tabular}
        \caption{\textbf{Ablation study on few shot scoring.}}
        \label{tab:ablation-few-shot}
    \end{subtable}
    \hfill
    \begin{subtable}[t]
    {0.54\textwidth}  % 右侧子表格
        \renewcommand\tabcolsep{1pt}
        \centering
        \begin{tabular}{c|cccccc}
            \hline

            \hline

            \hline

            \hline
            
            & \multicolumn{5}{c}{\underline{\textit{\textbf{Video-Condition Alignment}}}} & \multirow{3}*{\textbf{Avg.}}\\

            Chain of &Video-text & Object-class & Color & Action & Scene \\
             Query & Consist. & Consist. & Consist. & Consist. & Consist.\\
            \hline
            
            \hline
              &$0.671$ &$0.690$ &$0.699$ & $0.662$&$0.675$ &$0.679$\\
             \checkmark&$\textbf{0.732}$&$\textbf{0.735}$ & $\textbf{0.750}$&$\textbf{0.718}$ &$\textbf{0.733}$ & $\textbf{0.7336}$\\
             \hline

            \hline
                
            \hline
                
            \hline
        \end{tabular}
        \caption{\textbf{Ablation Study on chain of query.}}
        \label{tab:ablation-coq}
    \end{subtable}

    \label{tab:ablation-two-tables}
\end{table*}
\section{Experiment}
\label{sec:exp}
\subsection{Video Generation Models}
We evaluate 4 open-source t2v models and 3 commercial models, spanning a range of advancements from earlier to more recent models. Specifically, we select 4 open-source models: LaVie~\citep{wang2023lavie}, 
Show-1~\citep{zhang2023show}, 
VideoCrafter2~\citep{chen2024videocrafter2}, and CogVideoX
\footnote{CogVideoX-5B,the larger model in the CogVideoX series. }~\citep{yang2024cogvideox}, as well as 4 commercial models: Pika-Beta~\citep{pikalab}. 
Kling~\citep{klingkuaishou}, Gen3~\citep{runwaygen3}, and  this diverse selection allows for a comprehensive evaluation of video generation techniques across different development stages, providing insights into the evolution of model performance.

\subsection{Human preference annotation}
We recruit 10 human annotators to manually rate each video\footnote{The human annotators are provided with an extensive guide to ensure they provide quality evaluations of the videos given the prompts. The detailed guide is included in the Appendix.}. For each generated video, we collect 4 human evaluation scores. The generated videos are shuffled and randomly assign to the pool of 10 evaluators. This leaves us a total of 35,196 evaluations. Both human and MLLM evaluations follow the same scale and guidelines. A human expert examines the score generated by human evaluators to gate-keep the quality of the evaluations. Table ~\ref{tab:agreement} shows the inter-rater alignment score produced by human rators, which is comparable to the human self-alignment scores produced by other studies~\citep{peng2024dreambench++}.

\subsection{Implementation details}
Our experiment adopts GPT-4o (Version: \texttt{gpt-4o-2024-08-06}) and GPT-4o-mini (Version: \texttt{gpt-4o-mini}) as MLLM and LLM, respectively. GPT-4o handles multimodal inputs including texts and video frames, while GPT-4o-mini is only prompted with texts. The evaluation instructions are written by humans, including task description, important notes, strategy, and output format. The full prompt of the evaluation process is included in the Appendix.

\section{Main results}
% In this section, we analyze the effectiveness, alignment, stability and robustness of the proposed benchmark through the following analysis.
% \subsection{Does \textbf{\textit{Video-Bench}} results align with existing benchmarks? 
% \textcolor{blue}{Is Video-Bench an Effective Indicator of Community Evolution?}}
\subsection{Comparison with existing evaluation methods}
As shown in Table~\ref{tab:human_align}, we compare evaluation methods from previous benchmarks (\textit{e.g.}, EvalCrafter~\citep{liu2023evalcrafter}, VBench~\citep{huang2024vbench}, and ComBench~\citep{sun2024t2v}) on our prompt suite, calculating Spearman correlations with human ratings. Our MLLM-based evaluation achieves the highest correlation. Notably, ComBench~\citep{sun2024t2v} uses single-round video description and evaluation, while our Chain of Query mechanism enables multi-round interactions, improving Video-Condition Alignment by $0.093$. This highlights our framework’s ability to capture richer visual information and enhance cross-modal semantic comparisons. Overall, Video-Bench has stronger correlations to metric-based evaluation over other LLM-based methods across all dimensions.

\subsection{Human preference alignment}
As reported by Table~\ref{tab:agreement}, its agreement with human scores is on par with the inter-rater agreement among humans, with an average score of $0.52$. To compare the distribution of the two evaluations on a larger sample, we calculate the mean difference after bootstrapping $1000$ iterations over $100k$ score pairs sampled with replacement. The average absolute mean difference across dimensions between our proposed approach and human evaluations is $0.18$. \textbf{\textit{This shows Video-Bench's potential to replicate human judgments.}} We include the results of mean difference and its $99\%$ confidence interval across dimensions in the Appendix.

\begin{table*}[htbp]
\caption{
\textbf{Evaluation results on different base models.} For each dimension, we randomly select 30 prompts for comparison.
}

\scriptsize
\renewcommand\tabcolsep{3.5pt}
\renewcommand\arraystretch{1.2}
\renewcommand{\footnote}{\fnsymbol{footnote}} 
\small
\setlength{\abovecaptionskip}{0.0cm}
\setlength{\belowcaptionskip}{-0.45cm}
\centering
\hspace*{0.00000000000000001cm}
% \scalebox{0.86}{%
\begin{tabular}{l|cccc|ccccc|c}

\hline

\hline

\hline

\hline

& \multicolumn{4}{c|}{\underline{\textit{\textbf{Video quality}}}} & \multicolumn{5}{c|}{\underline{\textit{\textbf{Video-Condition Alignment}}}} & \multirow{3}*{\textbf{Avg.}} \\

Base models&Imaging  & Aesthetic & Temporal & Motion & Video-text & Object-class & Color & Action & Scene \\
& Quality & Quality & Consist. & Effects & Consist. & Consist. & Consist. & Consist. & Consist. \\
\hline

\hline

Gemini1.5pro & $0.602$ & $0.583$ &$0.367$ & $0.340$ & $0.600$ & $0.656$& $0.602$ & $0.491$ & $0.579$ & $0.536$ \\
Qwen2vl-72b & $0.586$ & $0.51$ & $0.535$& $0.353$&$0.576$ &
 $0.669$ & $0.634$ &  $0.637$ & $0.600$ & $0567$ \\
GPT-4o-0513 & $0.711$ & $\textbf{0.690}$ & $0.484$& $0.427$&$0.657$ & $0.755$&$0.621$ & $0.621$ & $0.619$ & $0.621$ \\
GPT-4o-0806 & $\textbf{0.807}$ & $0.667$ & $0.494$& $\textbf{0.469}$& $\textbf{0.750}$ & $\textbf{0.767}$&$\textbf{0.676}$ & $\textbf{0.761}$ &$ \textbf{0.73}$ & $\textbf{0.680}$ \\
GPT-4o-1120 & $0.724$ & $0.651$ & $\textbf{0.538}$& $0.309$&$0.711$ & $0.749$&$0.621$ & $0.674$ & $0.619$ & $0.622$\\

\hline

\hline

\hline

\hline

\end{tabular}
% }
 \vspace{0.25cm}

\label{tab:ablation_on_base_model}
\end{table*}
\subsection{MLLM evaluation \textit{vs.} human evaluation}
% [Case study to prove that it has potential to surpass humans]}}
\label{surpass}
Table~\ref{tab:agreement} shows that human evaluations have low agreement on semantic consistency-related dimensions, i.e. the five dimensions under Video-Condition Alignment. Nevertheless, this corresponds to the findings from EvalCrafter~\citep{liu2023evalcrafter} that users tend to prioritize visual appeal over text-to-video alignment. 

Due to individual differences among human evaluators (some may give high scores to videos with good visual effects, even if they don't meet the requirements), human consistency tends to be lower in semantics-related evaluations~\citep{marr2010vision}, while the results from the proposed strategy do not exhibit such trend. In fact, when incorporating evaluations from Video-Bench, the agreement improves in the Video-Text Consistency dimension. This demonstrates that \textbf{\textit{our MLLM-based evaluation mitigates the perception bias commonly found in human assessments}}.

%\textbf{The advantage over human judgment.}
% Studies in visual neuroscience~\citep{marr2010vision} have shown that the brain processes visual information hierarchically. Lower-level visual areas detect simple features, while higher-level areas are responsible for integrating these features into objects and associating semantic meanings. This means that human evaluators that are more instinctive can bias towards better-looking videos but with lower alignment quality, whereas more meticulous evaluators would assign a lower score to those videos. As a result, the agreement level of such semantics-related metrics can be low as compared to metrics related to simple visual features. 

% \textbf{\textit{Video-Bench addresses the perception bias common in human evaluations, where visual aspects are often prioritized over semantics. Leveraging MLLMs' strengths in prompts-following, Video-Bench can ensure a comprehensive evaluation focused on semantic consistency.}} In fact, when including evaluations from Video-Bench, the agreement degree increases in Video-text Consistency dimension.

\subsection{Does MLLM generate stable evaluations? }
% [Stability experiments to prove that it's reliable]}}

Research~\citep{atil2024llm} has shown that no LLMs deliver repeatable outcomes across different tasks in different benchmarks. If there is significant variance across identical runs, it reduces the reliability and validity of the benchmark. To this end, we analyze the stability of our benchmark to make sure the evaluations are stable and the benchmark is reliable. We run the experiments on Imaging Quality dimension three times with identical and deterministic configuration. 

The results achieve a TARa@3 (Total Agreement Rate-answer across 3 runs) score of $0.67$, meaning that $67\%$ of the videos obtain the same rating from three repeated runs. In addition, to measure the scale of rating difference across the three runs, we calculate the Krippendorff's $\alpha$ as well, which achieves $0.867$. This means that even though only $67\%$ of the videos achieve the exact same rating, the agreement of the ratings across identical runs are is highly substantial. \textbf{\textit{The results show that Video-Bench produces evaluations with high agreement across different runs.}}
\subsection{{Robustness against small variations}}

A common issue found in traditional metric-based visual evaluations is that adding small perturbations hardly invisible to the human eye can easily fool the metrics~\citep{videoprocessing2024}. To measure the robustness of our method against such small variations, we apply Gaussian blur to a subset of our videos and re-run our evaluation method. 
Under the video-text consistency dimension, we observe less than $5\%$ relative percentage error.
This showcases \textbf{\textit{small variations in the input video that do not affect human judgment will not undermine MLLM evaluation either}}.
\subsection{Comparing \textit{v.s.} rating paradigm}
In video generation evaluations, both comparison-based~\citep{huang2024vbench} and rating-based~\citep{li2024genai} evaluations have been adopted by researchers. A popular type of comparison-based evaluation is arenas~\citep{huggingface_leaderboard}, where human evaluators pick the winner out of two choices. However, studies have found that both LLM~\citep{zheng2023judging} and humans~\citep{blunch1984position} suffer from position bias in comparison-based evaluations, where both tend to favor the first item they see. Another major drawback of pairwise comparison is its high cost. To obtain the relative ranking across $N$ models, $N$ evaluations is needed for rating-based approach while $N\times (N-1)/2$ evaluations are needed for comparison-based approach. The quadratic complexity bottlenecks the benchmark's capability in evaluating a large number of models. On the other hand, rating-based approaches may fail to capture the subtle differences between videos, especially in the quality related metrics. To mitigate this problem, we propose a few-shot scoring component where the evaluation model can have a sense of the relative quality across models. Table~\ref{tab:ablation-few-shot} shows that with few-shot scoring, the correlation with respect to human evaluations on average arises by $10.33\%$ across different dimensions without increasing runtime complexity. This shows that the proposed rating paradigm can achieve \textbf{\textit{high alignment with human evaluations}} while still being \textbf{\textit{low-cost and free of position bias}}.
\subsection{Ablation study}
Table~\ref{tab:ablation-two-tables} shows the ablation study on alignment with humans. We observe that our proposed components are all necessarily effective in reaching higher alignment with humans. Adding each component leads to a significant increase in the human-alignment score across all dimensions. Based on such observations,
we validate that both components are effective in aligning with human preference. Experiments on Table~\ref{tab:few_shot} confirm that more reference videos lead to better performance.

\subsection{Comparision on different base models}
Table~\ref{tab:ablation_on_base_model} indicates GPT-4o generally achieves superior video quality and alignment scores compared to Gemini1.5pro and Qwen2vl-72b, particularly in Imaging Quality ($0.807$) and Video-text Consistency (0.750) in GPT-4o-0806. However, performance does not consistently improve with newer GPT-4o versions. For instance, GPT-4o-1120 shows decreased Motion Effects (0.309 vs. 0.469 in GPT-4o-0806), suggesting potential regressions in temporal-motion detection across updates. Note that our benchmark results are recorded using the optimal version.

\subsection{Simple \textit{v.s.} complex prompt}
As shown in Table~\ref{tab:complex_prompts}, we tested state-of-the-arts (\textit{i.e.}, Gen3, Kling and Pika) on both simple and complex prompts (\textit{e.g.}, MovieGenBench), showing consistent performance across varying prompt lengths and complexities, demonstrating its robustness and versatility.

\begin{table}[t]
\raggedright  
\setlength{\tabcolsep}{3pt}  % 适当缩小列间距
\renewcommand\arraystretch{1.3}
\caption{\textbf{Performance on simple prompts \textit{vs.} complex prompts.}}
\label{tab:complex_prompts}  % 让表格可以在正文中引用
\resizebox{\columnwidth}{!}{  % 让表格适应单栏
    \begin{tabular}{l|cc|cc}
        \hline
        
        \hline

        \hline
        
        \hline

        \multirow{3}*{Prompt type} & \multicolumn{2}{c|}{\underline{\textit{\textbf{Video quality}}}} & \multicolumn{2}{c}{\underline{\textit{\textbf{Video-Condition Alignment}}}} \\
        & Imaging & Motion & Video-text & Action \\
        & Quality & Effects & Consistency & Consistency \\
        \hline
        
        \hline
        Simple prompt & $0.796$ & $0.475$ & $\textbf{0.749}$&$0.701$ \\
        Complex prompt & $\textbf{0.797}$ & $\textbf{0.484}$ & $0.725$ & $\textbf{0.704}$\\
        \hline
        
        \hline
        
        \hline

        \hline
    \end{tabular}
}
\end{table}

\begin{table}[t]
\raggedright  
\setlength{\tabcolsep}{3pt}  % 适当缩小列间距
\caption{\textbf{Video quality comparison under different $N$-shot settings.}}
\label{tab:few_shot}  % 让表格可以在正文中引用

\resizebox{\columnwidth}{!}{  % 让表格适应单栏
    \begin{tabular}{l|cccccc}
        \hline
        
        \hline

        \hline
        
        \hline

        \multirow{3}*{Num.} & \multicolumn{5}{c}{\underline{\textit{\textbf{Video quality}}}} \\
        & Imaging & Aesthetic & Temporal & Motion & Avg \\
        & Quality & Quality & Consistency & Effects & Score \\
        \hline
        
        \hline
1 & $0.461$ & $0.341$ & $0.456$ & $0.359$ & $0.404$ \\
3 & $0.596$ & $0.496$ & $\textbf{0.529}$ & $0.424$ & $0.511$ \\
5 & $0.611$ & $\textbf{0.511}$ & $\textbf{0.529}$ & $0.416$ & $0.517$ \\
7 & $\textbf{0.624}$ & $0.498$ & $0.498$ & $\textbf{0.557}$ & $\textbf{0.545}$ \\
        \hline
        
        \hline
        
        \hline

        \hline
    \end{tabular}
}
\end{table}

% \input{tables/VBench}
 % \small \bibliographystyle{ieeenat_fullname} \bibliography{main}

\section{Conclusion}
This paper has introduced Video-Bench, a human-aligned video generation benchmark leveraging evaluations from Multi-Modal Large Language Models (MLLMs). Extensive experiments and a human alignment study have demonstrated its advantages in efficiency , strong alignment with human preferences. In addition, we have provided insights into component design, emphasizing the potential for improving automatic evaluations through few-shot and chain-of-query technologies. Our work aims to support future research on video generation model development by providing a highly human-aligned benchmark using MLLMs in visual evaluation.
{\small
\bibliographystyle{ieeenat_fullname}
% \bibliography{cited}}

\clearpage
\appendix
\clearpage
\maketitlesupplementary
In Sec.~\ref{sec:more_details_on_evaluation_dimension}, we elaborate on our evaluation dimensions with comprehensive explanations and examples. Sec.~\ref{sec:additional_experimental_results} presents extended experimental results and analysis. Sec.~\ref{sec:potential_societal_impacts} discusses the broader implications and potential impacts on society. Finally, Sec.~\ref{sec:limitations_and_future_work} examines the current limitations and outlines promising directions for future research. 
%%%%% to be changed in the arxiv version %%%%%
\section{More Details on Evaluation Dimension}
\label{sec:more_details_on_evaluation_dimension}

\subsection{Video-Condition Consistency}
\subsubsection{Object Class Consistency}
\paragraph{Definition and scope}
Object class consistency evaluation assesses \textbf{\textit{the consistency between objects in the video and those specified in the text prompt}}. The assessment should consider the following key aspects:
\begin{itemize}
    \item \textbf{Generation accuracy}: Whether objects mentioned in the text are correctly generated.
    \item \textbf{Class identification}: Whether object categories are clearly identifiable.
    \item \textbf{Appearance fidelity}: Whether generated objects' appearance and structure align with objective reality and human perceptual expectations.
    \item \textbf{Deformation}: Whether objects maintain their structural integrity during motion.
\end{itemize}

\paragraph{Scoring criteria}
\begin{itemize}
    \item \blackcircle{1} \textbf{Poor consistency}: Objects are completely unrecognizable or fail to match the specified objects in the prompt.
    
    \item \blackcircle{2} \textbf{Moderate consistency}: Objects are barely recognizable as the specified class but exhibit one or more of the following issues:
    \begin{itemize}
        \item Partial object generation (\textit{e.g.}, only a hand visible when a complete person is specified).
        \item Feature mixing between specified object and other object classes.
        \item Unstable object characteristics (\textit{e.g.}, facial features appearing and disappearing).
        \item Presence of unspecified objects of the same category or multiple similar objects occupying significant space (\textit{e.g.}, a motorcycle consistently appearing alongside a specified car).
    \end{itemize}
    
    \item \blackcircle{3} \textbf{Good consistency}: Object classes remain correct and consistent throughout the entire video, avoiding all issues mentioned in the moderate consistency category.
\end{itemize}
\begin{figure}[h!]
    \centering
    \small
    Prompt: \textit{``A surfboard.''}
    \begin{minipage}{\columnwidth}
        \begin{subfigure}{\columnwidth}
        \centering
        \includegraphics[width=\columnwidth]{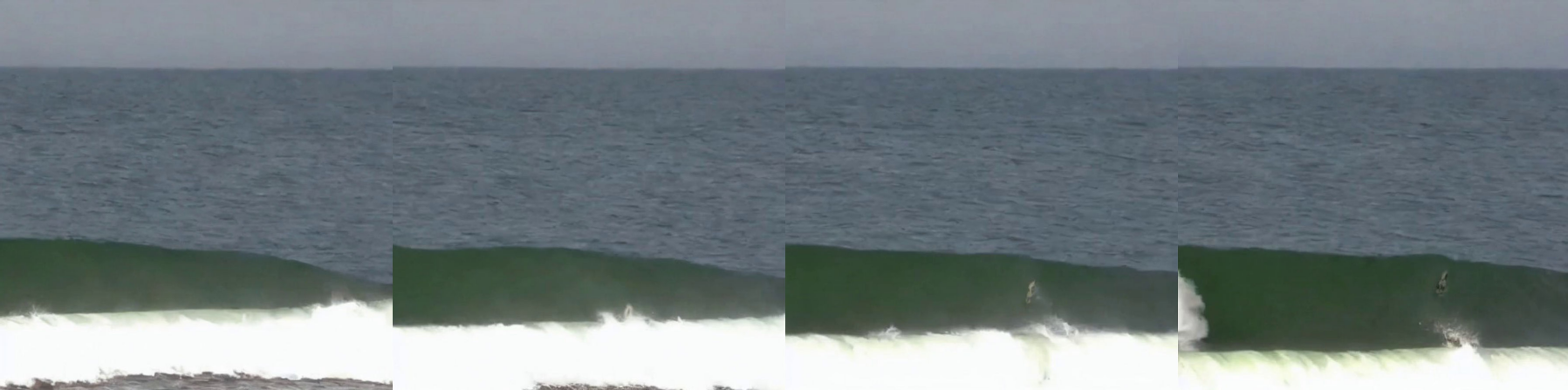} % replace with your image file
        \caption{Poor object class consistency (Score=1)}
        \label{fig:object_1}
    \end{subfigure}

        \begin{subfigure}{\columnwidth}
        \centering
        \includegraphics[width=\columnwidth]{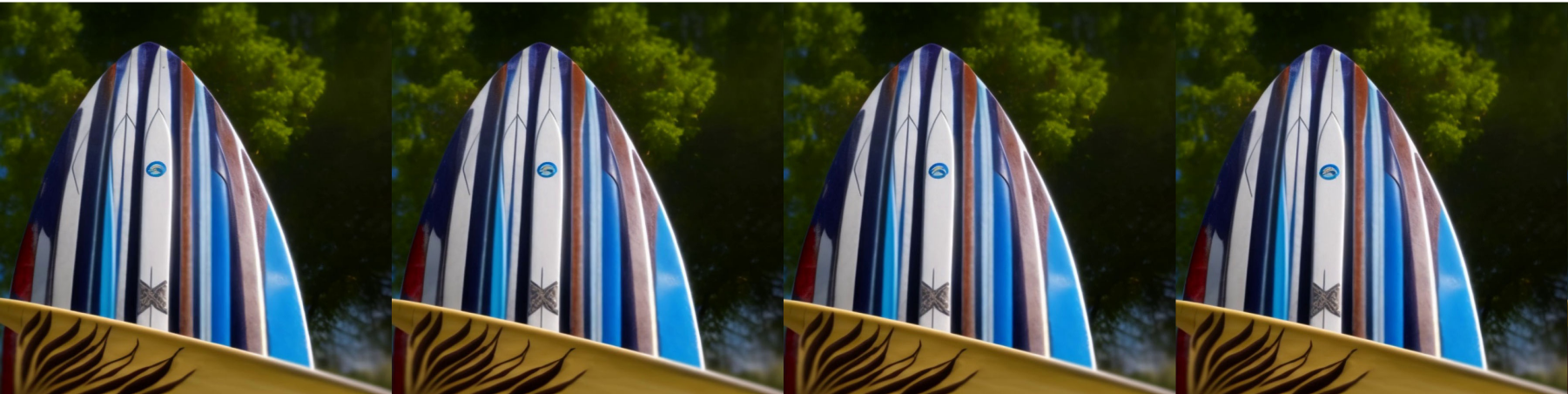} % replace with your image file
        \caption{Moderate object class consistency (Score=2)}
        \label{fig:object_2}
    \end{subfigure}
    
    \begin{subfigure}{\columnwidth}
        \centering
        \includegraphics[width=\columnwidth]{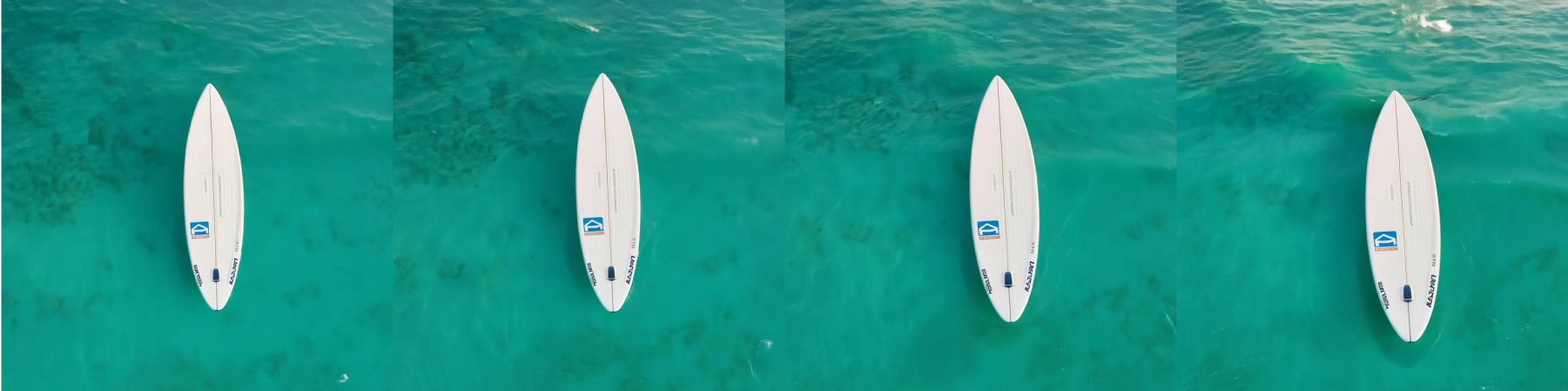} % replace with your image file
        \caption{Good object class consistency (Score=3)}
        \label{fig:object_3}
    \end{subfigure}
    \end{minipage}

    \caption{
    \textbf{Comparative examples of object class consistency assessment.} 
    \textbf{(a) Poor}: Generated scene shows only ocean waves without any surfboard, completely failing to meet the prompt requirement.
    \textbf{(b) Moderate}: Multiple surfboards present but with additional decorative elements and patterns that complicate the straightforward prompt requirement.
    \textbf{(c) Good}: Clean white surfboard rendered consistently throughout the sequence, precisely matching the simple prompt specification.
    }
    \label{fig:object_cases}
\end{figure}

\subsubsection{Action Consistency}
\paragraph{Definition and scope}
Action consistency evaluation assesses \textbf{\textit{the consistency between actions in the video and those specified in the text prompt}}. The assessment should consider the following key aspects:
\begin{itemize}
    \item \textbf{Generation accuracy}: Whether actions mentioned in the text are correctly generated.
    \item \textbf{Action identification}: Whether actions are clearly identifiable.
    \item \textbf{Process fidelity}: Whether the appearance and progression of actions align with objective reality and human perceptual expectations.
\end{itemize}

\paragraph{Scoring criteria}
\begin{itemize}
    \item \blackcircle{1} \textbf{Poor consistency}: Actions are either completely unrecognizable or incorrectly generated.
    
    \item \blackcircle{2} \textbf{Moderate consistency}: Actions are partially consistent but exhibit one or more of the following issues:
    \begin{itemize}
        \item Significant deviation from the realistic appearance or progression of the action.
        \item Incomplete action representation, either in terms of viewpoint or temporal coverage, showing only a fragment of the complete action.
    \end{itemize}
    
    \item \blackcircle{3} \textbf{Good consistency}: Actions fully align with the prompt specifications and avoid all issues mentioned in the moderate consistency category.
\end{itemize}

\paragraph{Important notes}
\begin{itemize}
    \item This metric focuses primarily on the presence and accuracy of actions in the video rather than their dynamic presentation or motion effects. However, completely static videos that fail to show any movement should be scored as inconsistent with objective understanding.
\end{itemize}
\begin{figure}[h!]
    \centering
    \small
    Prompt: ``\textit{A person is marching.}''
    \begin{minipage}{\columnwidth}
        \begin{subfigure}{\columnwidth}
        \centering
        \includegraphics[width=\columnwidth]{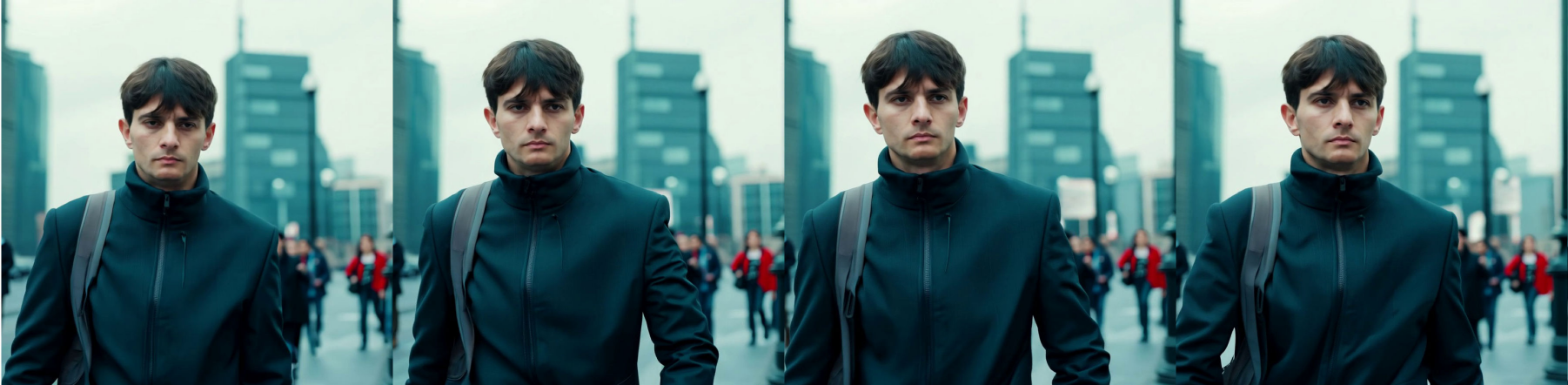} % replace with your image file
        \caption{Poor action consistency (Score=1)}
        \label{fig:action_1}
    \end{subfigure}
        \begin{subfigure}{\columnwidth}
        \centering
        \includegraphics[width=\columnwidth]{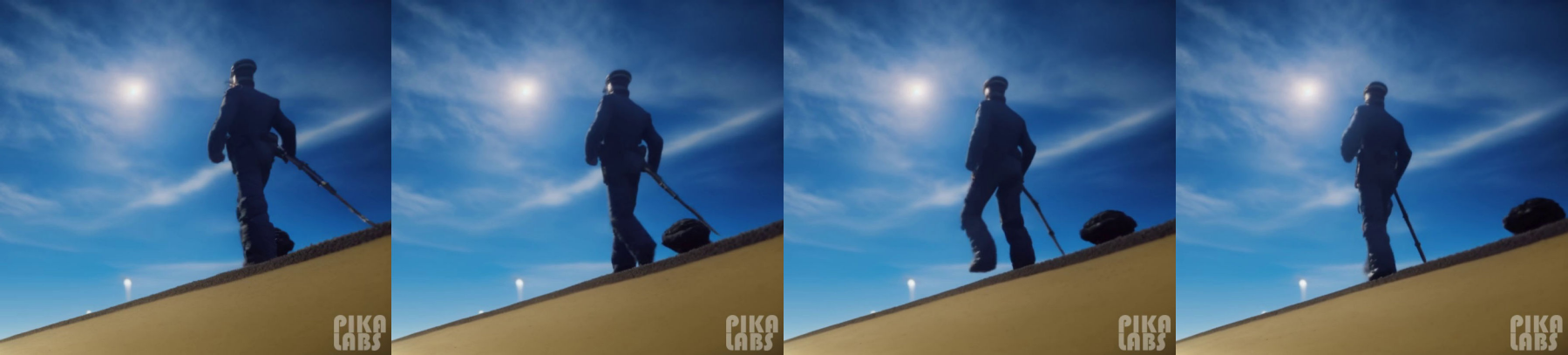} % replace with your image file
        \caption{Moderate action consistency (Score=2)}
        \label{fig:action_2}
    \end{subfigure}
    \begin{subfigure}{\columnwidth}
        \centering
        \includegraphics[width=\columnwidth]{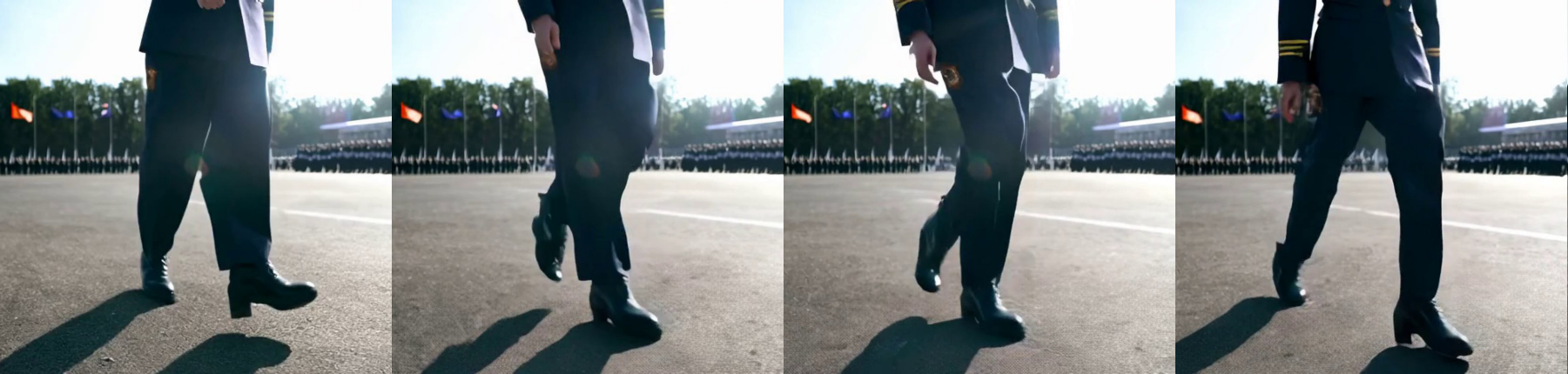} % replace with your image file
        \caption{Good action consistency (Score=3)}
        \label{fig:action_3}
    \end{subfigure}
    \end{minipage}

    \caption{
    \textbf{Comparative examples of action consistency assessment.}
    \textbf{(a) Poor}: The man appears to be walking naturally in the video rather than marching, failing to demonstrate any marching motion, contradicting the prompted action completely.
\textbf{(b) Moderate}: The figure shows walking/sliding motion, but the movement appears unnatural and doesn't fully capture the rhythmic, structured nature of marching.
\textbf{(c) Good}: Clear marching action with proper leg movement and posture, displaying fluid and natural progression of steps that precisely matches the prompted action.
    }
    \label{fig:action_cases}
\end{figure}

\subsubsection{Color Consistency}
\paragraph{Definition and scope}
Color consistency evaluation assesses \textbf{\textit{the degree of color matching between the video and the provided text prompt}}. The assessment should consider the following key aspects:
\begin{itemize}
    \item \textbf{Color consistency}: Whether colors align with the text prompt and maintain stability throughout the video without sudden changes.
    \item \textbf{Color placement}: Whether colors appear on the correct objects or within appropriate scenes.
\end{itemize}

\paragraph{Scoring criteria}
\begin{itemize}
    \item \blackcircle{1} \textbf{Poor consistency}: Objects are either incorrectly generated or display colors that completely deviate from the text prompt specifications.
    
    \item \blackcircle{2} \textbf{Moderate consistency}: Correct colors appear in the video but exhibit imperfections in one or more of the following aspects:
    \begin{itemize}
        \item Incorrect color allocation (\textit{e.g.}, colors appearing in background instead of on intended objects).
        \item Color instability with sudden changes or variations in object coloring.
        \item Color confusion where objects display correct colors mixed with significant areas of unintended colors (\textit{e.g.}, a requested white vase generated as black and white).
        \item Poor color distinction between objects and background.
        \item Approximate color matching within the same spectrum but lacking precision (\textit{e.g.}, pink versus purple, yellow versus orange).
    \end{itemize}
    
    \item \blackcircle{3} \textbf{Good consistency}: Colors demonstrate high fidelity to the text prompt, maintain stability throughout the video, show correct distribution, and exhibit no sudden changes or inconsistencies. The work avoids all issues mentioned in the moderate consistency category.
\end{itemize}
\begin{figure}[h!]
    \centering
    \small
    Prompt: ``\textit{A yellow cat.}''
    \begin{minipage}{\columnwidth}
        \begin{subfigure}{\columnwidth}
        \centering
        \includegraphics[width=\columnwidth]{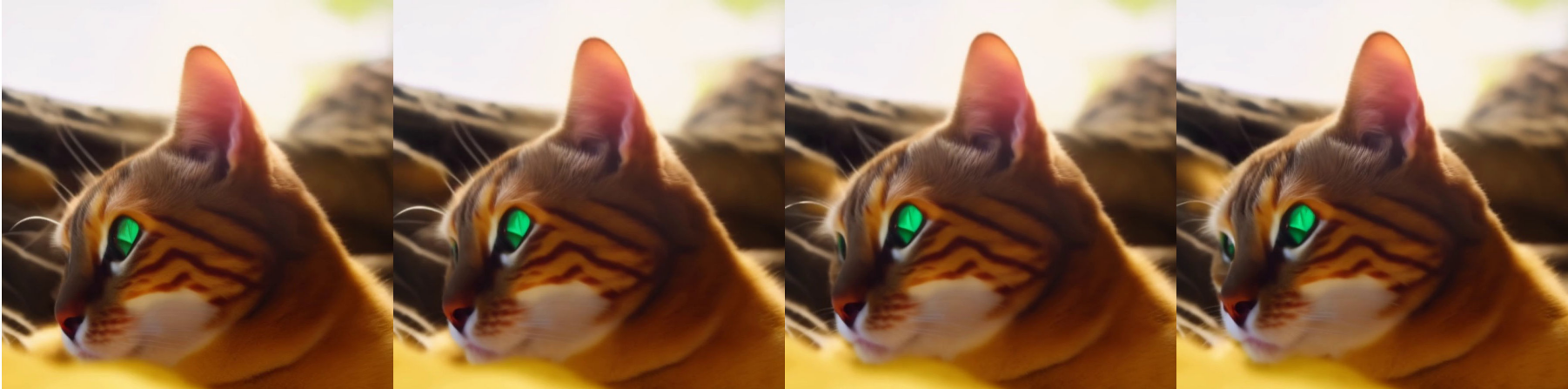} % replace with your image file
        \caption{Poor color consistency (Score=1)}
        \label{fig:color_1}
    \end{subfigure}
        % \vspace{0.5cm}  % 添加垂直间距，调整图像间的间隔
    \begin{subfigure}{\columnwidth}
        \centering
        \includegraphics[width=\columnwidth]{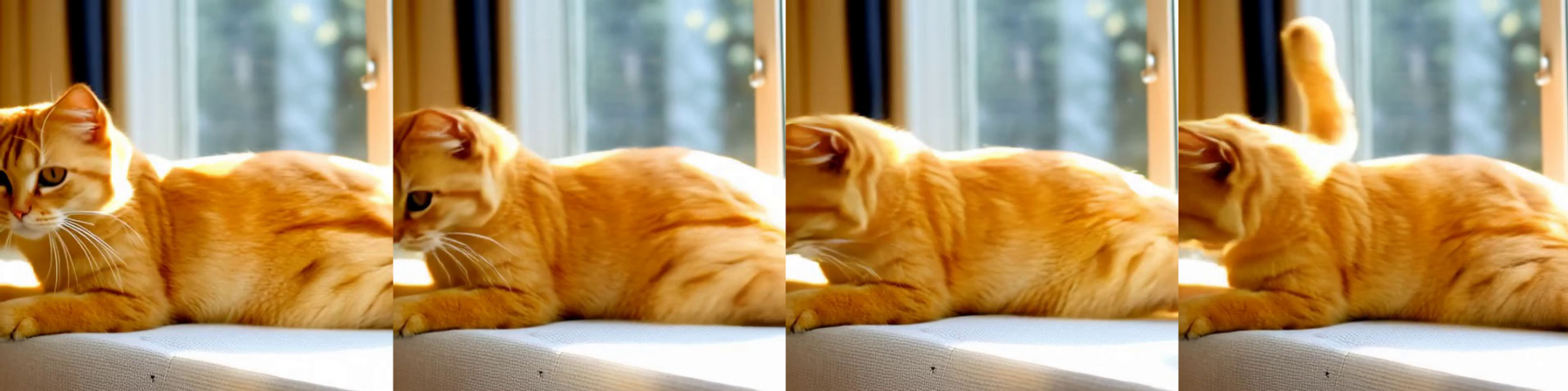} % replace with your image file
        \caption{Moderate color consistency (Score=2)}
        \label{fig:color_2}
    \end{subfigure}
    \begin{subfigure}{\columnwidth}
        \centering
        \includegraphics[width=\columnwidth]{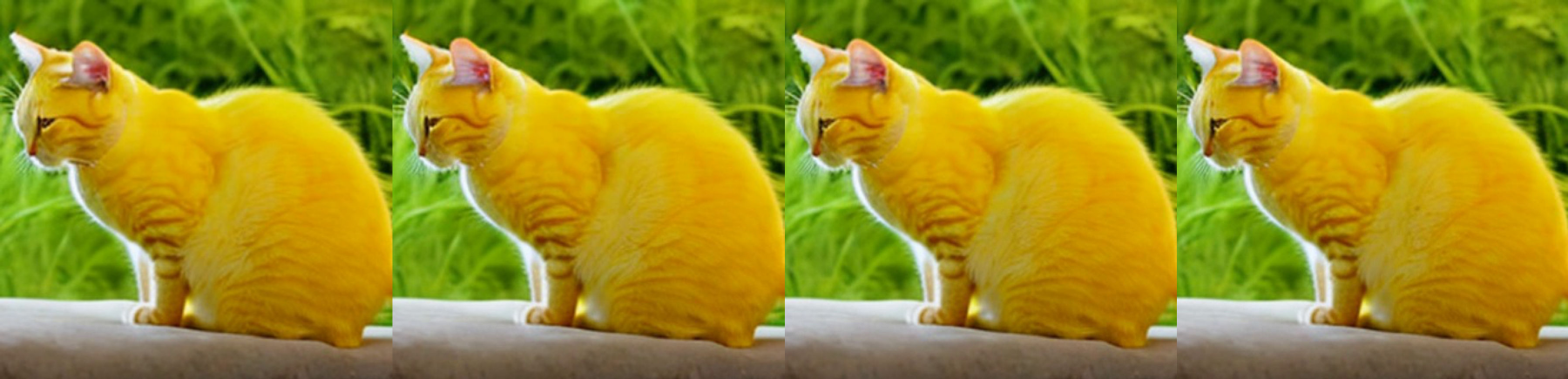} % replace with your image file
        \caption{Good color consistency (Score=3)}
        \label{fig:color_3}
    \end{subfigure}
    \end{minipage}

    \caption{
    \textbf{Comparative examples of color consistency assessment.} 
    \textbf{(a) Poor}: The generated cat appears brown/orange with unnatural blue-green eyes, significantly deviating from the prompted yellow color.
    \textbf{(b) Moderate}: Cat displays an orange/ginger coloration that, while consistent throughout the sequence, falls within a similar but distinct color spectrum from the requested yellow.
    \textbf{(c) Good}: The cat exhibits pure yellow coloring that precisely matches the prompt specification, maintaining a consistent hue throughout the sequence.
}
    \label{fig:color_cases}
\end{figure}

\subsubsection{Scene Consistency}
\paragraph{Definition and scope}
Scene Consistency evaluation assesses \textbf{\textit{the consistency between scenes in the video and those specified in the text prompt}}. The assessment should consider the following key aspects:
\begin{itemize}
    \item \textbf{Generation accuracy}: Whether scenes mentioned in the text are correctly generated.
    \item \textbf{Scene identification}: Whether scenes are clearly identifiable.
    \item \textbf{Element fidelity}: Whether the appearance and structure of scene elements align with objective reality and human perceptual expectations.
\end{itemize}

\paragraph{Scoring criteria}
\begin{itemize}
    \item \blackcircle{1} \textbf{Poor consistency}: Scene generation is completely unrelated to the text prompt and scenes are difficult to identify.
    
    \item \blackcircle{2} \textbf{Moderate consistency}: Scenes are barely recognizable and exhibit one or more of the following issues:
    \begin{itemize}
        \item Partial scene generation without showing the complete scene context.
        \item Display of limited scene characteristics (\textit{e.g.}, only bread in a bakery, only a sink in a bathroom).
        \item Scene generation that is similar but not precisely matching the specified scene.
    \end{itemize}
    
    \item \blackcircle{3} \textbf{Good consistency}: Scenes are clearly identifiable and align with human subjective understanding of objective world arrangements.
\end{itemize}

\paragraph{Important notes}
\begin{itemize}
    \item For ambiguous scene terms, scoring should use the most comprehensive interpretation among the generated results as the standard. For example, if ``bathroom'' is generated as a complete bathroom with a bathtub by one model and as a simple washroom with only a mirror, sink, or toilet by another, the complete bathroom setting should be used as the reference standard.
\end{itemize}
\begin{figure}[h!]
    \centering
    \small
    Prompt: ``\textit{Arch.}''
    \begin{minipage}{\columnwidth}
        \begin{subfigure}{\columnwidth}
        \centering
        \includegraphics[width=\columnwidth]{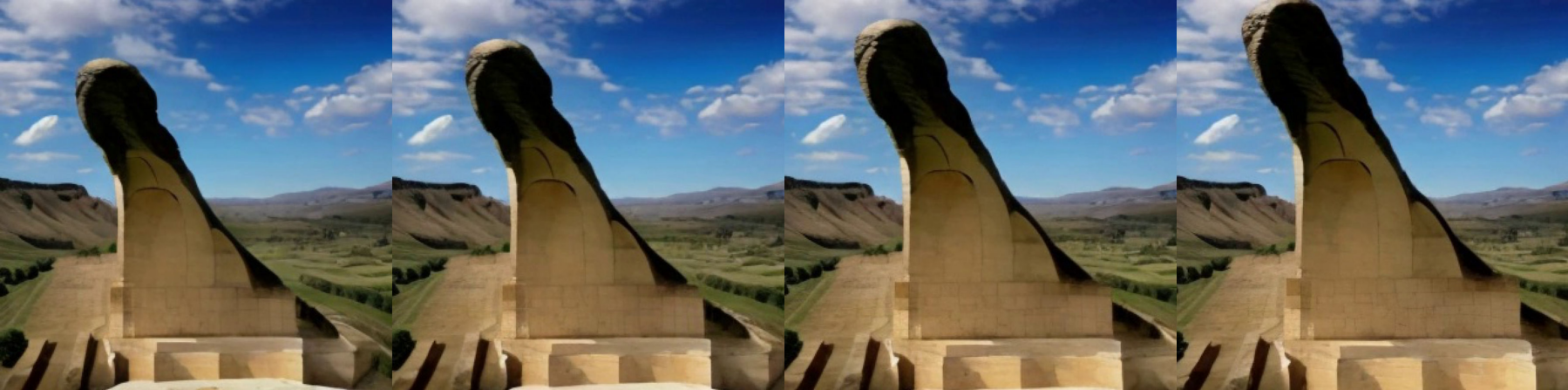} % replace with your image file
        \caption{Poor scene consistency (Score=1)}
        \label{fig:scene_1}
    \end{subfigure}
    \begin{subfigure}{\columnwidth}
        \centering
        \includegraphics[width=\columnwidth]{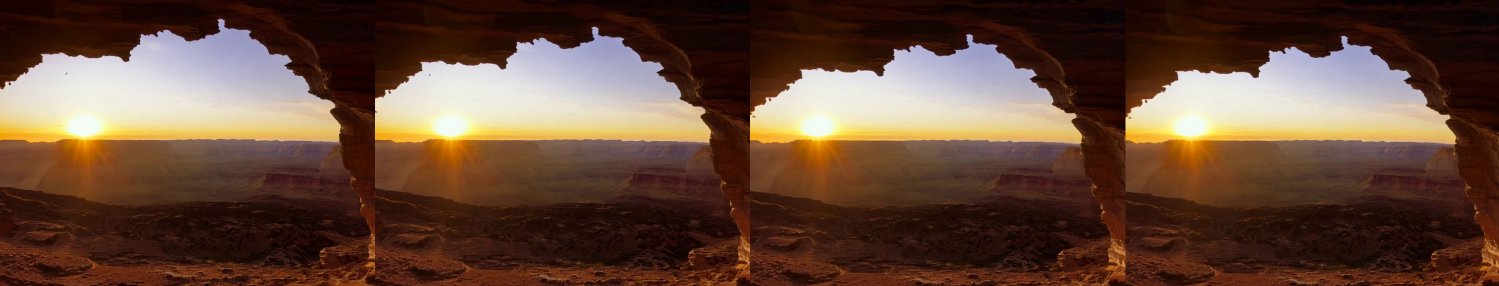} % replace with your image file
        \caption{Moderate scene consistency (Score=2)}
        \label{fig:scene_2}
    \end{subfigure}
    \begin{subfigure}{\columnwidth}
        \centering
        \includegraphics[width=\columnwidth]{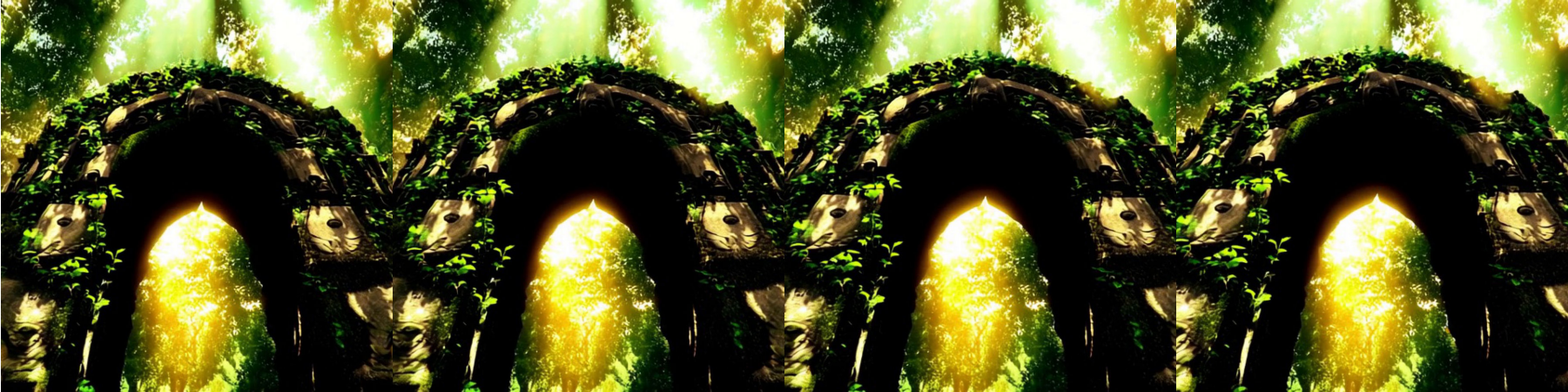} % replace with your image file
        \caption{Good scene consistency (Score=3)}
        \label{fig:scene_3}
    \end{subfigure}
    \end{minipage}

    \caption{
\textbf{Comparative examples of scene consistency assessment.}
\textbf{(a) Poor}: The scene displays an architectural structure but lacks clear arch definition, with distracting foreground elements and inconsistent structural representation.
\textbf{(b) Moderate}: The sunset scene through an arch frame demonstrates a recognizable arch structure, though the dramatic lighting and silhouette effect partially obscure architectural details.
\textbf{(c) Good}: Garden arch with natural foliage shows a clear, well-defined arch structure maintained consistently throughout the sequence, with proper architectural form and depth.
}
    \label{fig:scene_cases}
\end{figure}

\subsubsection{Video-text Consistency}
\paragraph{Definition and scope}
Video-text consistency evaluates \textbf{\textit{the comprehensive alignment between the video and the text prompt}}. The assessment should consider the following key aspects:
\begin{itemize}
    \item \textbf{Core element coverage}: Whether the video demonstrates all core elements mentioned in the text prompt (including humans, animals, actions, objects, scenes, styles, spatial relationships, and quantitative relationships).
    \item \textbf{Visual clarity}: Whether the video's image quality affects comprehension of its content.
\end{itemize}

\paragraph{Scoring criteria}
\begin{itemize}
    \item \blackcircle{1} \textbf{Very poor consistency}: Missing half or more of the key elements, demonstrating very weak consistency, or visual quality so poor that video comprehension is impossible.
    
    \item \blackcircle{2} \textbf{Poor consistency}: Video includes most key elements but most are insufficiently generated, or visual quality is inadequate for determining consistency with the text prompt.
    
    \item \blackcircle{3} \textbf{Moderate consistency}: Video either includes most key elements with sufficient generation or includes all elements but most are insufficiently generated. Visual quality is adequate for determining consistency with the text prompt.
    
    \item \blackcircle{4} \textbf{Good consistency}: Video includes all key elements, with some elements insufficiently generated. Visual quality is adequate for determining consistency with the text prompt.
    
    \item \blackcircle{5} \textbf{Excellent consistency}: Video includes all key elements with sufficient generation and complete alignment with the text prompt. Visual quality is adequate for determining consistency with the text prompt.
\end{itemize}

\paragraph{Important notes}
\begin{itemize}
    \item Insufficient generation refers to elements that are present but fail to meet consistency requirements, such as low visibility in actions or objects that don't conform to objective world appearances.
    \item ``Most'' is determined by the number of key elements in the prompt, typically not exceeding 5 elements, thus ``most'' generally means $N-1$ elements.
    \item This metric does not have high requirements for visual quality. Superior visual quality is not a prerequisite for high scores.
\end{itemize}
\begin{figure}[htp]
    \centering
    \small
    Prompt: ``\textit{Two pandas discussing an academic paper.}''
    \begin{minipage}{\columnwidth}

    \begin{subfigure}{\columnwidth}
        \centering
        \includegraphics[width=\columnwidth]{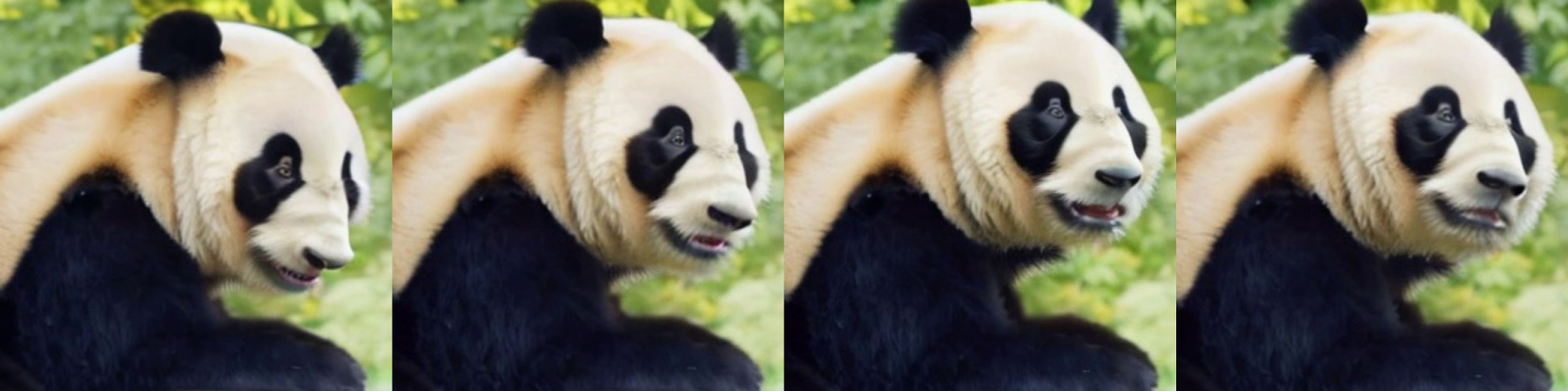} % replace with your image file
        \caption{Very poor video-text consistency (Score=2)}
        \label{fig:overall_1}
    \end{subfigure}
    \end{minipage}

    \begin{subfigure}{\columnwidth}
        \centering
        \includegraphics[width=\columnwidth]{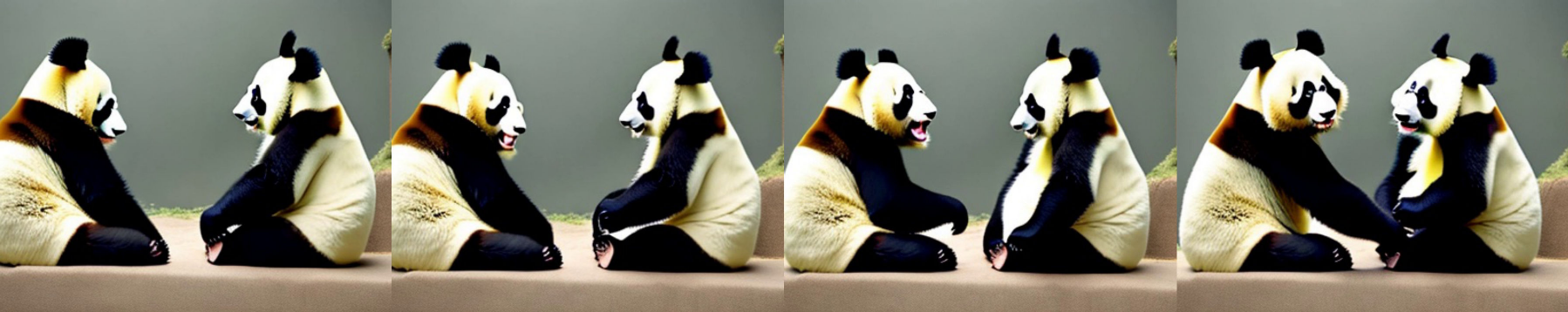} % replace with your image file
        \caption{Moderate video-text consistency (Score=3)}
        \label{fig:overall_2}
    \end{subfigure}
    
    \begin{subfigure}{\columnwidth}
        \centering
        \includegraphics[width=\columnwidth]{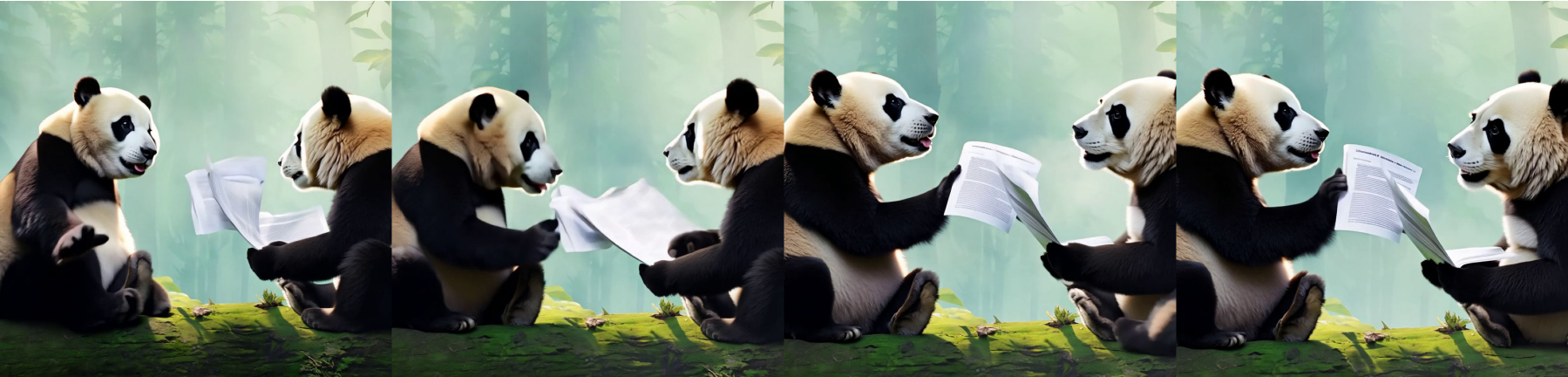} % replace with your image file
        \caption{Excellent video-text consistency (Score=5)}
        \label{fig:overall_3}
    \end{subfigure}
    \caption{\textbf{Comparative examples of video-text consistency assessment.} 
\textbf{(a) Very poor}: Shows only a single panda with no interaction or academic elements, significantly deviating from the prompt requirements.
\textbf{(b) Moderate}: While two pandas are shown and appear to be facing each other, the interaction lacks clear indication of academic discussion or the presence of a paper.
\textbf{(c) Excellent}: The scene perfectly captures the prompt, showing two pandas in an interactive pose with what appears to be a paper between them, complete with a natural academic discussion setting.
    }
    \label{fig:overall_cases}
\end{figure}

\subsection{Video Quality}
\subsubsection{Imaging Quality}
Imaging quality evaluates \textbf{\textit{the visual fidelity and clarity of the generated video compared to standard high-definition content}}. The assessment should consider the following key aspects:
\begin{itemize}
\item \textbf{Image clarity}: Overall sharpness, resolution, and detail preservation throughout the video.
\item \textbf{Visual artifacts}: Presence of noise, distortion, overexposure, or other technical imperfections.
\end{itemize}

\paragraph{Scoring criteria}
\begin{itemize}
\item \blackcircle{1} \textbf{Very poor quality}: Severe visual artifacts with obvious distortions, extreme blurriness, excessive noise, and significant overexposure issues that severely impact the viewing experience.

\item \blackcircle{2} \textbf{Poor quality}: Notable visual artifacts with apparent distortions, general blurriness, and noise that detract from the natural appearance and viewing experience.

\item \blackcircle{3} \textbf{Moderate quality}: Resolution comparable to 480p standard definition, with minor artifacts, slight noise, and occasional exposure issues that moderately affect the viewing experience.

\item \blackcircle{4} \textbf{Good quality}: Resolution comparable to 720p high definition, with minimal artifacts and generally pleasant viewing experience.

\item \blackcircle{5} \textbf{Excellent quality}: Resolution comparable to 1080p full HD or better, with no discernible artifacts, providing an exceptional viewing experience comparable to professional video content.
\end{itemize}
\begin{figure}[htp]
    \centering
    \begin{minipage}{\columnwidth}
    \begin{subfigure}{\columnwidth}
        \centering
        \small
        Prompt: ``\textit{A person is jogging.}''
        \includegraphics[width=\columnwidth]{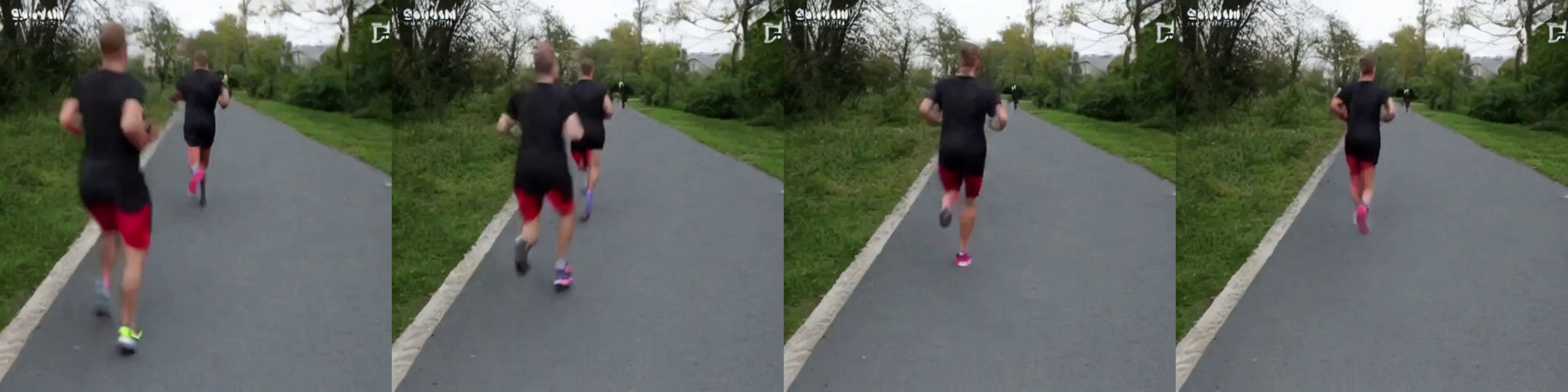} % replace with your image file
     \caption{Very poor imaging quality (Score=1)}
        \label{fig:imaging_2}
    \end{subfigure}

        \begin{subfigure}{\columnwidth}
        \centering
    
        \includegraphics[width=\columnwidth]{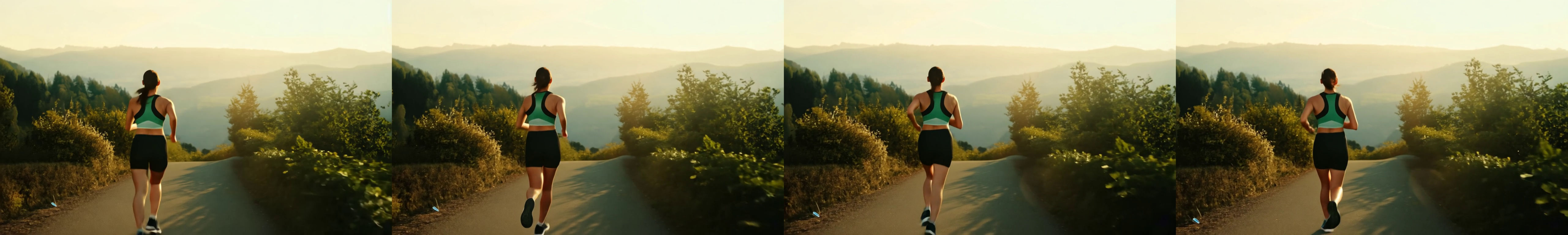} % replace with your image file
     \caption{Excellent imaging quality (Score=5)}
        \label{fig:imaging_5}
    \end{subfigure}
    \end{minipage}
    \caption{\textbf{Comparative examples of imaging quality assessment.} \textbf{(a) Very poor}: The generated video exhibits poor video clarity with visible noise, unstable camera movement, and inconsistent lighting conditions. \textbf{(b) Excellent}: It demonstrates superior visual fidelity with stable framing, natural lighting, sharp details, and professional cinematographic quality.}
    \label{fig:imaging_examples}
\end{figure}

\subsubsection{Aesthetic Quality}
\paragraph{Definition and scope} 
Aesthetic quality evaluation encompasses \textbf{\textit{the artistic and compositional elements of video production}}, including structural arrangement, color utilization, compositional effectiveness, visual appeal, and overall harmonic integration. 
The assessment of video-text consistency should consider the following key aspects:
\begin{itemize}
    \item \textbf{Structural coherence}: Whether the arrangement and composition of subjects (people or objects) in the video are logically sound and aesthetically pleasing, rather than causing psychological discomfort.
    \item \textbf{Color application}: The appropriateness and effectiveness of color usage throughout the video sequence.
    \item \textbf{Compositional efficacy}: Whether the composition effectively captures and presents all necessary information specified in the text prompt.
    \item \textbf{Visual appeal}: The video's capacity to maintain visual engagement.
    \item \textbf{Overall harmony}: The degree to which all elements work together cohesively to create a unified and harmonious visual experience.
\end{itemize}

\paragraph{Scoring criteria}
\begin{itemize}
    \item \blackcircle{1} \textbf{Very poor aesthetic quality}: The work exhibits severe deficiencies in color utilization, composition, and clarity. It lacks visual appeal and emotional expression, with poor overall harmonic integration.
    \item \blackcircle{2} \textbf{Poor aesthetic quality}: The work demonstrates notable deficiencies in specific aspects, such as discordant color schemes or inadequate composition, significantly compromising the overall aesthetic experience.
    \item \blackcircle{3} \textbf{Moderate aesthetic quality}: The work shows average performance across most dimensions, with possible minor deficiencies in certain aspects, while maintaining a basic aesthetic experience.
    \item \blackcircle{4} \textbf{Good aesthetic quality}: The work demonstrates strong execution in color usage, composition, and clarity, delivering a satisfying visual experience with appropriate emotional expression and creative elements.
    \item \blackcircle{5} \textbf{Excellent aesthetic quality}: The work excels in all aspects, achieving high standards in color utilization, composition, and clarity. It delivers powerful visual impact and profound emotional expression, providing an outstanding aesthetic experience.
\end{itemize}
\begin{figure}[htp]
    \centering
    \begin{minipage}{\columnwidth}
    \begin{subfigure}{\columnwidth}
        \centering
        \small
        Prompt: ``\textit{A bear is climbing trees.}''
        \includegraphics[width=\columnwidth]{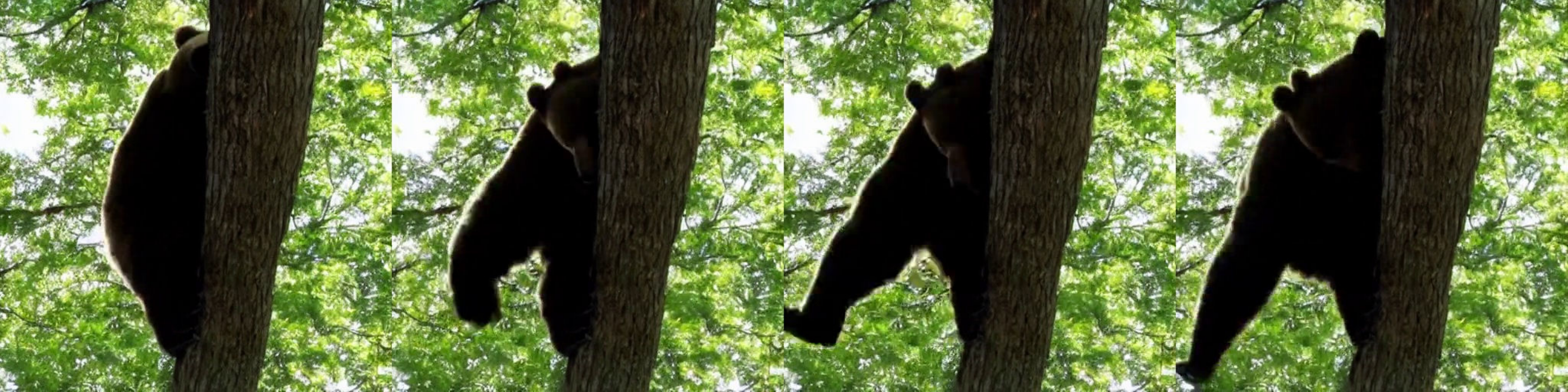} % replace with your image file
     \caption{Very poor aesthetic quality (Score=1)}
        \label{fig:aesthetic_1}
    \end{subfigure}

    \begin{subfigure}{\columnwidth}
        \centering
        \includegraphics[width=\columnwidth]{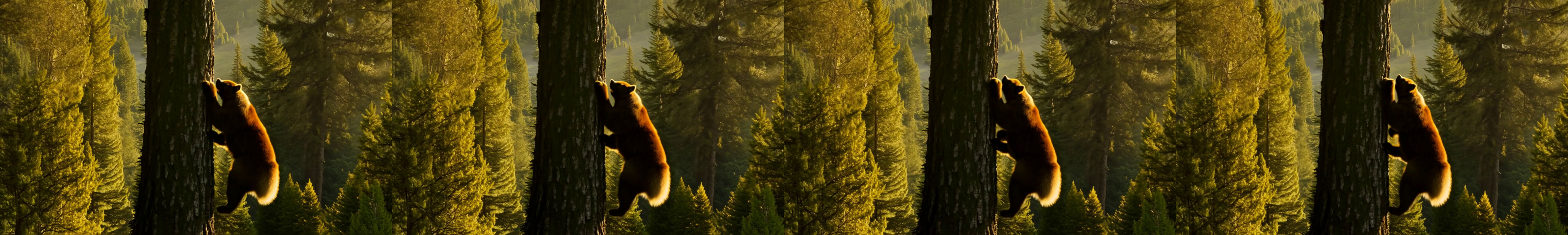} % replace with your image file
     \caption{Excellent aesthetic quality (Score=5)}
        \label{fig:aesthetic_5}
    \end{subfigure}

    \end{minipage}
    \caption{\textbf{Comparative examples of aesthetic quality assessment.} \textbf{(a) Very poor}: Bear climbing tree sequence with flat composition, lacking visual depth and artistic consideration in framing and lighting.
\textbf{(b) Excellent}: Cat leaping sequence captured in golden-hour lighting with atmospheric forest backdrop, demonstrating sophisticated composition and cinematic appeal.}
    \label{fig:aesthetic_examples}
\end{figure}

\subsubsection{Temporal Consistency}
\paragraph{Definition and scope}
Temporal consistency evaluates \textbf{\textit{the consistency of semantic and visual features between consecutive frames}} in the video, ensuring smooth transitions without abrupt changes or unnatural jumps. The assessment encompasses two primary aspects:
\begin{itemize}
    \item \textbf{Visual feature consistency}:
    \begin{itemize}
        \item \textbf{Color and brightness}: Smooth transitions between consecutive frames without flickering or sudden changes in illumination.
        \item \textbf{Texture and detail}: Maintenance of consistent object textures and details across frames without unexpected blur or clarity shifts.
    \end{itemize}
    
    \item \textbf{Semantic consistency}:
    \begin{itemize}
        \item \textbf{Object position and shape}: Preservation of consistent object positioning and morphology between frames without unnatural deformation or displacement.
        \item \textbf{Scene layout}: Maintenance of consistent scene composition and background elements across frames.
        \item \textbf{Subject coherence}: Stability of main subjects across consecutive frames without abrupt changes or unnatural transitions.
    \end{itemize}
\end{itemize}

\paragraph{Scoring criteria}
\begin{itemize}
    \item \blackcircle{1} \textbf{Very poor consistency}: Significant inconsistencies in color, brightness, and texture between frames with obvious flickering or sudden changes. Semantic features show discrepancies in object positioning and scene layout, with main subjects exhibiting sudden or unnatural variations.
    
    \item \blackcircle{2} \textbf{Poor consistency}: Notable inconsistencies in visual features. Semantic features maintain general consistency but occasionally display issues with object positioning and scene layout. Main subjects may exhibit minor inconsistencies.
    
    \item \blackcircle{3} \textbf{Moderate consistency}: Visual features typically maintain consistency with minor fluctuations in color, brightness, and texture. Semantic features show general consistency with slight issues affecting object position, shape, and scene layout coherence. Main subjects maintain general consistency with minor deviations.
    
    \item \blackcircle{4} \textbf{Good consistency}: Visual features maintain consistency between frames with smooth transitions in color, brightness, and texture. Semantic features demonstrate coherence with stable object positions, shapes, and scene layout. Main subjects maintain consistency with only minor inconsistencies that don't significantly impact overall coherence.
    
    \item \blackcircle{5} \textbf{Excellent consistency}: All visual features demonstrate seamless consistency between frames without perceptible flickering or sudden changes. Semantic features show complete consistency in object positions, shapes, scene layout, and background. Main subjects maintain perfect consistency without notable deviations that would affect viewer perception of continuity.
\end{itemize}
\begin{figure}[h!]
    \centering
    \begin{minipage}{\columnwidth}
    \begin{subfigure}{\columnwidth}
        \centering
        \small
        Prompt: ``\textit{A person is skateboarding.}''
        \includegraphics[width=\columnwidth]{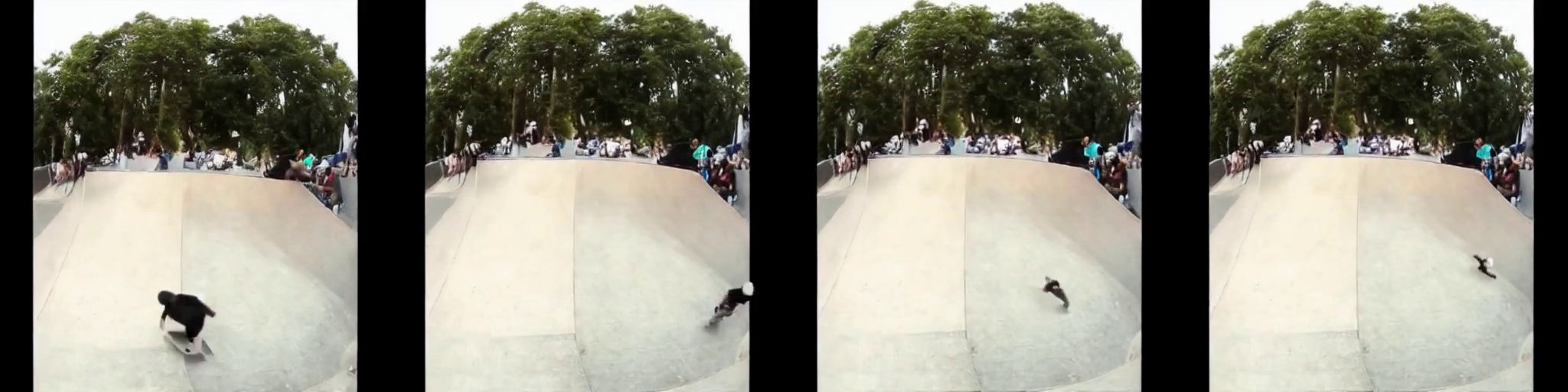} % replace with your image file
     \caption{Very poor temporal consistency (Score=1)}
        \label{fig:temporal_1}
    \end{subfigure}
    \begin{subfigure}{\columnwidth}
        \centering
        \includegraphics[width=\columnwidth]{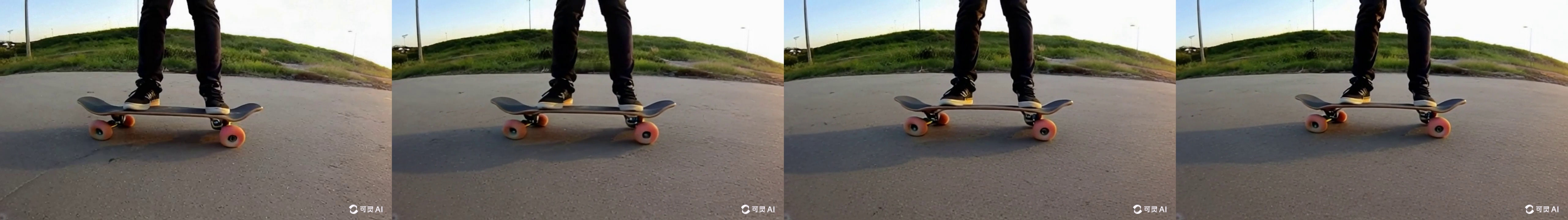} % replace with your image file
     \caption{Excellent temporal consistency (Score=5)}
        \label{fig:temporal_5}
    \end{subfigure}
    \end{minipage}
    \caption{\textbf{Comparative examples of temporal consistency assessment.}
    \textbf{(a) Very poor:} Skateboarding sequence shows inconsistent object placement and scaling between frames, with abrupt position shifts disrupting motion continuity.
    \textbf{(b) Excellent:} Skateboarding motion displays coherent object positioning and consistent spatial relationships across frames, maintaining seamless temporal flow.
    }
    \label{fig:Temporal_examples}
\end{figure}

\subsubsection{Motion Effects}
\paragraph{Definition and scope}
Motion effects evaluate \textbf{\textit{the quality of subject motion and its interaction with the environment}} in the video. The assessment should consider the following key aspects:
\begin{itemize}
    \item \textbf{Physical accuracy}: Whether object motion trajectories conform to physical laws such as inertia and gravity.
    \item \textbf{Dynamic blur}: Whether motion blur appropriately corresponds to the speed and direction of movement.
    \item \textbf{Environmental interaction}: Whether the relationship between moving objects and their background is coherent, including expected occlusion and reflections.
    \item \textbf{Lighting physics}: Whether changes in shadows and lighting during object movement align with physical laws, enhancing scene realism.
\end{itemize}

\paragraph{Scoring criteria}
\begin{itemize}
    \item \blackcircle{1} \textbf{Very poor effects}: Motion trajectories are severely incorrect, or the primary characteristics of movement are generated so poorly that the motion is barely recognizable. Clear violations of physical laws are present, and dynamic blur is either absent or completely misaligned with the motion.
    
    \item \blackcircle{2} \textbf{Poor effects}: Motion trajectories are poorly generated and movement is barely recognizable. Dynamic blur is inconsistent with movement speed and direction, and there are obvious issues with object interaction with background and lighting.
    
    \item \blackcircle{3} \textbf{Moderate effects}: Motion effects are generally present and movement is recognizable, but exhibits one of the following issues:
    \begin{itemize}
        \item Compromised motion smoothness, with noticeable frame-to-frame inconsistencies or abrupt changes disrupting motion fluidity.
        \item Inadequate or excessive dynamic blur application, failing to accurately reflect movement speed and direction.
        \item Partially maintained motion consistency, but certain elements like object-environment interaction or lighting changes lack convincing portrayal.
    \end{itemize}
    
    \item \blackcircle{4} \textbf{Good effects}: Movement is recognizable, motion trajectories and dynamic blur are mostly coherent, but certain aspects of motion appear unnatural and do not align with human subjective understanding of objective world changes.
    
    \item \blackcircle{5} \textbf{Excellent effects}: Movement is clearly recognizable, motion trajectories are accurate, dynamic blur is appropriately applied, and interaction of moving objects with their environment, including shadows and lighting, is seamlessly integrated and realistic.
\end{itemize}

\paragraph{Important notes}
\begin{itemize}
    \item This metric focuses specifically on dynamic presentation and effects, not on motion consistency with the text prompt. Consistency does not affect the scoring.
    \item In this metric, videos displaying static or no motion should be scored as 1 or 2.
\end{itemize}
\begin{figure}[htp]
    \centering
    \begin{minipage}{\columnwidth}
    \begin{subfigure}{\columnwidth}
        \centering
        \small
        Prompt: ``\textit{A person is catching or throwing baseball.}''
        \includegraphics[width=\columnwidth]{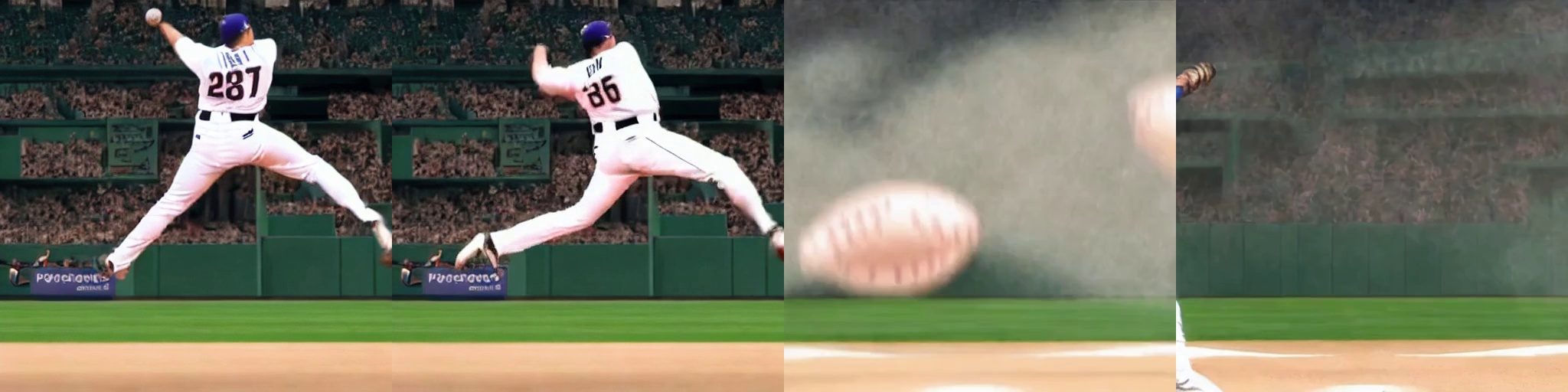} % replace with your image file
     \caption{Very poor motion Effects (Score=1)}
        \label{fig:motion_1}
    \end{subfigure}

    \begin{subfigure}{\columnwidth}
        \centering
        \includegraphics[width=\columnwidth]{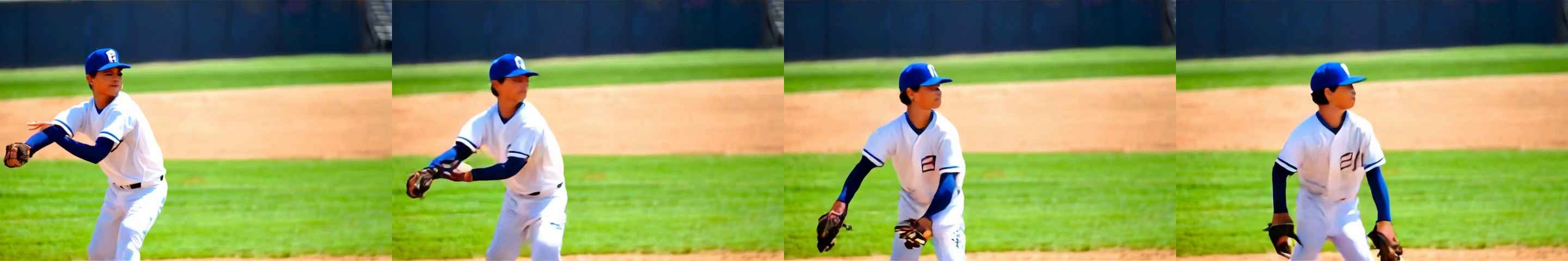} % replace with your image file
     \caption{Excellent motion Effects (Score=5)}
        \label{fig:motion_5}
    \end{subfigure}

    \end{minipage}

    \caption{
    \textbf{Comparative examples of motion effects assessment.}
    \textbf{(a) Very poor}: Pitcher motion sequence exhibits abrupt transitions with insufficient temporal continuity, resulting in unrealistic motion blur and motion artifacts.
\textbf{(b) Excellent}: Baseball player's pitching action rendered with smooth frame transitions and natural motion progression, maintaining consistent motion quality throughout the sequence.
    }
    \label{fig:Temporal_examples}
\end{figure}

\section{Additional Experimental Results}
\label{sec:additional_experimental_results}
\subsection{Statistical Analysis of Evaluation Discrepancy}
Table~\ref{tab:ci} presents a comprehensive statistical analysis of the evaluation discrepancies between our MLLM-based framework and human assessments across nine critical dimensions. The results reveal several noteworthy patterns:
\begin{itemize}
    \item \textit{\textbf{Video Quality Metrics:}} Among the four video quality metrics, temporal consistency shows the highest positive mean difference $(0.31)$, suggesting our framework maintains stricter standards in assessing temporal coherence. Interestingly, motion effects exhibit the only negative mean difference $(-0.26)$, indicating human evaluators are more sensitive to motion-related qualities. This contrast highlights the complementary nature of machine and human evaluation capabilities.
    \item \textit{\textbf{Alignment Metrics:}} In the video-condition alignment category, video-text consistency $(0.19)$ and action consistency $(0.22)$ demonstrate the most substantial positive differences. This aligns with our findings in the main paper, where the framework showed superior performance in detecting subtle semantic misalignments. The narrower confidence intervals in object-class consistency [$0.03, 0.05$] and color consistency [$0.04, 0.06$] suggest more stable and reliable evaluations in these aspects.
    \item \textit{\textbf{Statistical Significance:}} All confidence intervals at the $99\%$ level exclude zero, indicating statistically significant differences between MLLM and human evaluations across all dimensions. 
The tight confidence intervals, particularly in object-class and color consistency evaluations, demonstrate the robustness and reliability of our framework.
These findings quantitatively support our framework's capability to provide more stringent and consistent evaluations in most dimensions, while also revealing areas where human perception remains uniquely valuable (\textit{e.g.}, motion effects). 
\end{itemize}
\begin{table*}[htbp]
% \vspace{-0.6cm}
\caption{
\textbf{\textbf{\textit{Quick evaluation with our mini-split}}.} Higher scores indicate better performance. The best score in each dimension is highlighted in bold.
}

\scriptsize
\renewcommand\tabcolsep{2.6pt}
\renewcommand\arraystretch{1.2}
\renewcommand{\footnote}{\fnsymbol{footnote}} 
\small
\setlength{\abovecaptionskip}{0.0cm}
\setlength{\belowcaptionskip}{-0.45cm}
\centering
\hspace*{0.00000000000000001cm}
% \scalebox{0.86}{%
\resizebox{\textwidth}{!}{
\begin{tabular}{l|ccccc|cccccc|c}

\hline

\hline

\hline

\hline

\multirow{3}*{Model} & \multicolumn{5}{c|}{\underline{\textit{\textbf{Video quality}}}} & \multicolumn{6}{c|}{\underline{\textit{\textbf{Video-Condition Alignment}}}} & Overall\\

 & Imaging  & Aesthetic & Temporal & Motion & Avg & Video-text & Object-class & Color & Action & Scene & Avg& Avg\\
& Quality & Quality & Consist. & Effects & Rank& Consist. & Consist. & Consist. & Consist. & Consist. & Rank & Rank\\
\hline

\hline
Sora~\citep{yang2024cogvideox} &$\textbf{4.68}$ &$\textbf{4.64} $&$\textbf{4.96}$ & $4.24$& $\textbf{1.25}$&$ 4.48$&$2.88$&$2.92$ &$2.80$&$\textbf{2.96}$&$\textbf{2.20}$ &  $\textbf{1.78}$\\

Cogvideox~\citep{yang2024cogvideox} &$3.80$ &$3.96$ &$4.08$ &$3.84$& $4.00$&$\textbf{4.56}$&$2.80$&$2.84$&$\textbf{2.84}$&$2.92$&$ 2.80$ &  $3.30$\\

Gen3~\citep{runwaygen3} &$4.56$ & $4.56$  & $4.92$ &$\textbf{4.68}$& $1.75$&$4.36$&$\textbf{2.96}$&$2.80$&$2.56$&$2.88$& $3.80$&$2.89$\\

Kling~\citep{klingkuaishou} &  $4.16$ &$3.92$ &$4.40$ &$3.20$& $4.00$&$4.08$&$2.64$&$\textbf{2.96}$&$2.44$&$2.76$& $5.20$&$4.67$\\

VideoCrafter2~\citep{chen2024videocrafter2} & $4.00$&$4.00$  & $3.60$& $2.60$&$5.25$&$4.28$&$2.92$&$\textbf{2.96}$&$2.60$&$2.80$& $3.60$&$4.33$\\

LaVie~\citep{wang2023lavie} & $2.84$& $2.88$&$3.04$&$2.36$&$8.00$&$3.80$&$2.80$&$2.92$&$2.28$&$2.56$& $5.20$&$6.44$\\

PiKa-Beta~\citep{pikalab} &$3.60$ &$3.84$&$3.92$& $2.80$&$6.00 $&$3.80$&$2.40$&$2.76$&$2.68$&$2.72$&$ 7.40$&$6.78$\\

Show-1~\citep{zhang2023show} & $3.08$ &$3.24$ &$4.08$&$3.24$& $5.50$&$4.40$&$2.88$&$2.76$&$2.633$&$2.56$&$ 4.60$&$5.00$\\

\hline

\hline

\hline

\hline

\end{tabular}
}
% \vspace{0.25cm}

\label{tab:sora_test}
% \vspace{-0.2cm}
\end{table*}
\subsection{Mini-split for quick performance evaluation} 
To quickly evaluate video generation performance, we proposed the min-split scheme and conducted a comprehensive test on Sora~\ref{tab:sora_test}. The results show that Sora excels in video quality, motion effects, and video-condition alignment, outperforming Gen-3 and CogVideoX. However, there are some limitations in video-text consistency, temporal coherence, and motion generation, such as content mismatch, unnatural frame transitions, and inconsistent motion trajectories.

Specifically, we randomly selected 25 representative prompts (about one-third of the dataset) and generated 25 videos with Sora. The generation parameters were set to 720p resolution, 16:9 aspect ratio (to avoid wide-angle distortions from a 1:1 aspect ratio) and 5-second duration. Among the 9 evaluation metrics, Sora ranked first in 5 and demonstrated strong competitiveness across the remaining metrics, showcasing its potential as one of the most advanced video generation models available.

As shown in the results, Sora ranked first in 5 out of the 9 evaluated metrics and performed competitively across the remaining metrics. Notably, in all Video Quality metrics and in Video-Text Consistency, Sora matched or outperformed Gen-3 and CogVideoX, which had previously led these categories by a significant margin. Furthermore, Sora exhibited superior performance in Motion Effects and most Video-Condition Alignment metrics, except for Video-Text Consistency. Even under low-configuration settings (720p resolution and 5-second duration), Sora has demonstrated itself as one of the most advanced video generation models available.

However, case studies reveal certain limitations in Sora’s performance, particularly in video-text consistency, temporal frame coherence, and motion generation.

% Case 1: "A squirrel eating a burger" – Sora did not effectively generate the "eating a burger" action.

% Case 3: "A person is hula hooping" – While Sora generated all the necessary elements for the action, the overall motion sequence and execution were suboptimal.

% Case 4: "A person is motorcycling" – The generated video contained an unrealistic scene where the motorcycle made an abrupt, physically implausible turn in an extremely short time without any prior indication.

% Case 2: "An astronaut feeding ducks on a sunny afternoon, reflection from the water." – This case exemplifies multiple issues. While the video appeared visually clear and aesthetically appealing, the feeding motion was unnatural, and at a certain frame, the scene abruptly changed to a close-up of an astronaut’s face, further highlighting inconsistencies in motion and video-text alignment.

These findings suggest that while Sora is an impressive video generation model with greater stability compared to its predecessors, it has not yet reached the level of a true "world model." Its outputs still exhibit artifacts commonly associated with AI-generated content, indicating room for further improvement.

%通过实验和上述的结果讨论，我们验证了我们方法和人类偏好高度对齐，据此我们将其应用在最新发布的Sora视频生成模型上来评估这款‘世界模拟器’在视频生成上的表现，并借此为后续评估其他创新模型建立工作流程。

%首先我们从各指标中分别选取了25个(约占总数据集的1/3)代表性的prompt组成了quick test，使用Sora各生成1条视频共25条视频。生成参数分辨率720P,比例16:9\footnote{squre(1:1)会在部分动作上出现广角影响质量},时长5s；

%根据上表可以看到sora在9个指标的5个指标中取得了第一名，其他指标的表现也都在前列，并且在Video quality的所有指标和Video-text Consistency上，sora超越或保持齐平了在之前评估中分别断层领先的Gen3和CogVideoX。在Motion Effects以及除Video-text Consistency之外的其他Video-Condition Alignment指标上的表现也超越了绝大多数模型。仅使用了720p和5s的低配参数下Sora已经是当下最优秀的视频生成模型(之一)。

%但经过case study，发现sora在视频-文本一致性，时序帧间一致性以及动作生成上存在缺陷。
%Case1 A squirrel eating a burger 显然sora在生成”eating a burger“做的并不好
%Case3-A person is hula hooping 则表现出sora在生成动作时虽然把动作所需的元素生成完整但整个动作过程以及动作效果生成的并不理想。
%Case4-A person is motorcycling 是所生成视频的一小段，摩托车在毫无征兆的情况下在极短的时间内完成了一次不符合物理规律的原地转弯。
%Case2-An astronaut feeding ducks on a sunny afternoon, reflection from the water. 则是这些问题的集大成者，视频虽然非常清晰并且也很有意境(审美质量)，但不正常的喂食动作以及视频从某一帧突然变成一位宇航员的脸，都证明sora在上述三个问题上存在缺陷。

%（根据结果我建议往Sora目前看来是非常棒的视频生成模型，虽然比其他模型更稳定但并没有到world model的程度去写）还是很容易被识别成是AI生成的，还是存在于其他AI模型一样的一些缺陷。（这几句总结放在最后）
% \input{table/ablation}

\subsection{Comparison of long and short prompts}
% Table~\ref{tab:complex_prompts} presents a quick test on five representative video generation methods (\textit{i.e.}, Gen3, CogVideoX, Kling, VideoCrafter2, and LaVie) using both short (original) and complex prompts (\textit{e.g.}, MovieGenBench). The evaluation was conducted across four representative dimensions, with each dimension containing 25 prompts. The results indicate consistent performance across varying prompt lengths and complexities, demonstrating the robustness and versatility of these methods. 

The original prompts mostly contain at least one object, one action, and one scene. For example: \textit{``A fat rabbit wearing a purple robe walking through a fantasy landscape.''} Longer prompts build upon this by incorporating more objects, actions, and scenes. For example: \textit{``A couple sits at a peaceful lakeside picnic, occasionally reaching into a basket for food, while the gentle ripples on the lake reflect the shifting colors of the sky.''} It contains four subjects, two actions, one scene, and more complex semantic details. Even longer prompts may include: \textit{``A grandmother with neatly combed grey hair stands behind a colorful birthday cake with numerous candles at a wood dining room table, expression is one of pure joy and happiness, with a happy glow in her eye. She leans forward and blows out the candles with a gentle puff, the cake has pink frosting and sprinkles and the candles cease to flicker, the grandmother wears a light blue blouse adorned with floral patterns, several happy friends and family sitting at the table can be seen celebrating, out of focus. The scene is beautifully captured, cinematic, showing a 3/4 view of the grandmother and the dining room. Warm color tones and soft lighting enhance the mood.''}

% The complex prompts include those with an average length of \textcolor{red}{A couple sits at a peaceful lakeside picnic, occasionally reaching into a basket for food, while the gentle ripples on the lake reflect the shifting colors of the sky.}, while the longest prompt contains \textcolor{red}{A grandmother with neatly combed grey hair stands behind a colorful birthday cake with numerous candles at a wood dining room table, expression is one of pure joy and happiness, with a happy glow in her eye. She leans forward and blows out the candles with a gentle puff, the cake has pink frosting and sprinkles and the candles cease to flicker, the grandmother wears a light blue blouse adorned with floral patterns, several happy friends and family sitting at the table can be seen celebrating, out of focus. The scene is beautifully captured, cinematic, showing a 3/4 view of the grandmother and the dining room. Warm color tones and soft lighting enhance the mood.} 

% Table~\ref{tab:simple_vs_complex} tested state-of-the-arts (\textit{i.e.}, Gen3, Kling and Pika) on both short (original) and complex prompts (\textit{e.g.}, MovieGenBench), showing consistent performance across varying prompt lengths and complexities, demonstrating its robustness and versatility.
% \input{table/complex_prompts}

\subsection{Formulation of few-shot scoring}
All videos in a $N$-shot batch are input simultaneously, enabling comparison via 
$P(s_k | v_1,...,v_N, s_1,...,s_{(k-1)})$. 
As $k$ represents the index of the video, all videos $v$ are fully referenced.
% , whereas the score s can only refer to up to the $(k-1)$-th one.
% 
An in-batch video leverages the context of others and significantly outperforms direct scoring without any reference.
% First video ($k=1$) leverage the context of all video ($v_1$, ..., $v_N$), and significantly outperforms direct scoring without any reference, as shown in Table~\ref{tab:few_shot} ($N=1$ \textit{v.s.} $N > 1$).
% 
Inspired by in-context learning, our method groundedly adjusts scoring based on relative quality within the batch.
% 
% Experiments on Table~\ref{tab:few_shot} confirm that more reference videos lead to better performance, and a batch of size 5 is usually enough for producing best evaluation performance for multiple input videos.
%%%%%%%%% REFERENCES
% \input{table/fewshot_scoring}

\subsection{Can Machine Surpass Human in Video Evaluation?}
Our MLLM-based framework demonstrates superior discrimination ability over human evaluators in specialized content assessment. As shown in Figure~\ref{fig:outperform_cases}, MLLM consistently identifies subtle semantic distinctions that human evaluators often overlook. For instance, in object category evaluation (Figure~\ref{fig:outperform_cases} (a,b)), MLLM correctly differentiates between skiing and snowboarding, while humans mistakenly equate them. Similarly, in action assessment (Figure~\ref{fig:outperform_cases} (c,d)), MLLM precisely distinguishes air drumming from actual drum playing, a nuance that human evaluators miss. These cases reveal that MLLM's \textit{\textbf{comprehensive domain knowledge}} enables more rigorous semantic understanding, leading to more accurate and reliable video evaluation than human assessments.
\begin{figure}[htp]
    \centering
    \small
    Prompt: ``\textit{Skis}''
    \begin{minipage}{\columnwidth}
    \begin{subfigure}{\columnwidth}
        \centering
        \includegraphics[width=\columnwidth]{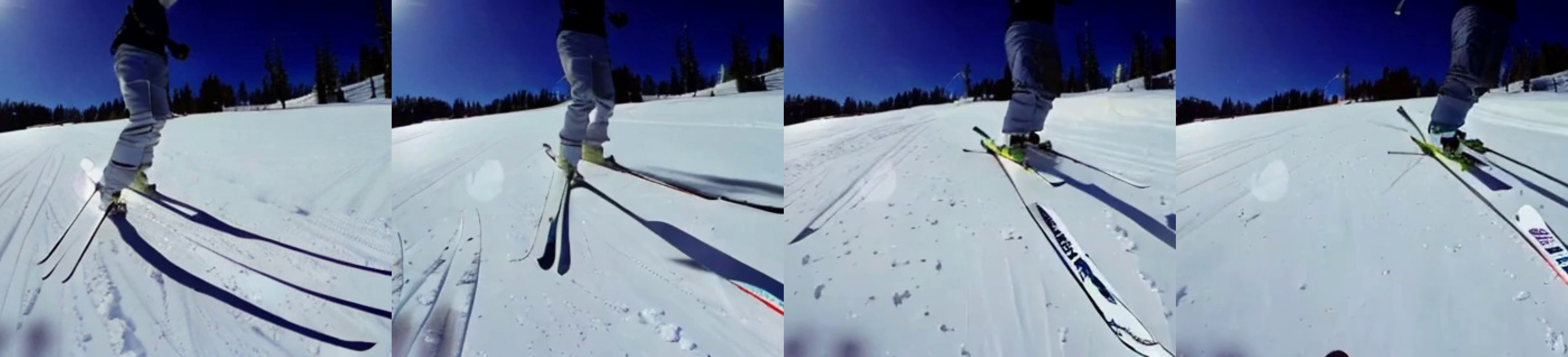} % replace with your image file
        \caption{Human: Good (Score=3) \cmark; MLLM: Good (Score=3) \cmark}
        \label{fig:skisframe}
    \end{subfigure}
    \begin{subfigure}{\columnwidth}
        \centering
        \includegraphics[width=\columnwidth]{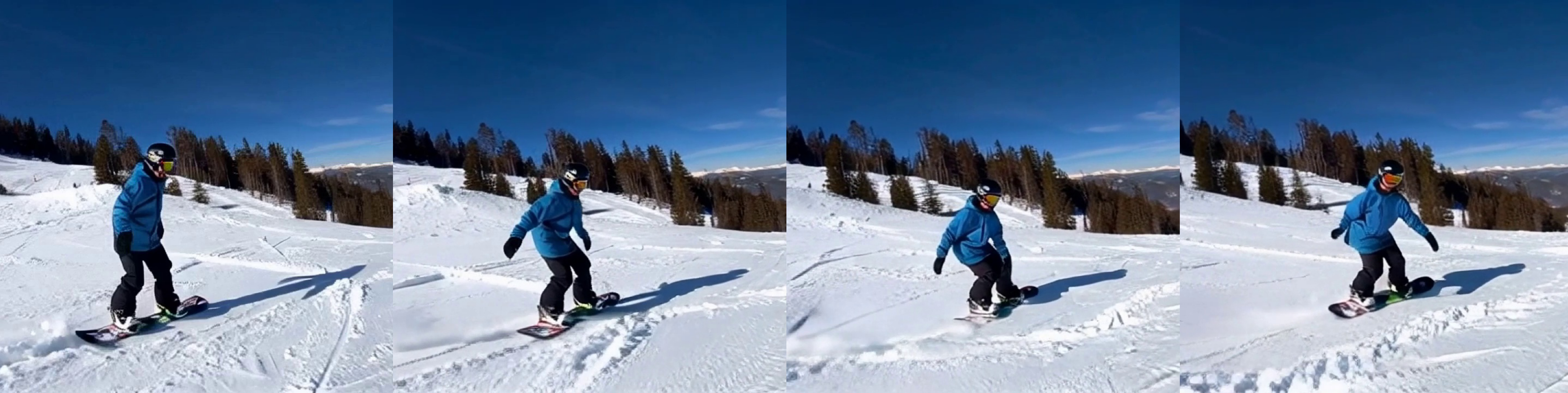} % replace with your image file
        \caption{Human: Good (Score=3) \xmark; MLLM: Moderate (Score=2) \cmark}
        \label{fig:snowboardframe}
    \end{subfigure}
    
    \vspace{0.5cm}
    \centering
    \small
    Prompt: ``\textit{A person is air drumming.}''
    \begin{subfigure}{\columnwidth}
        \centering
        \includegraphics[width=\columnwidth]{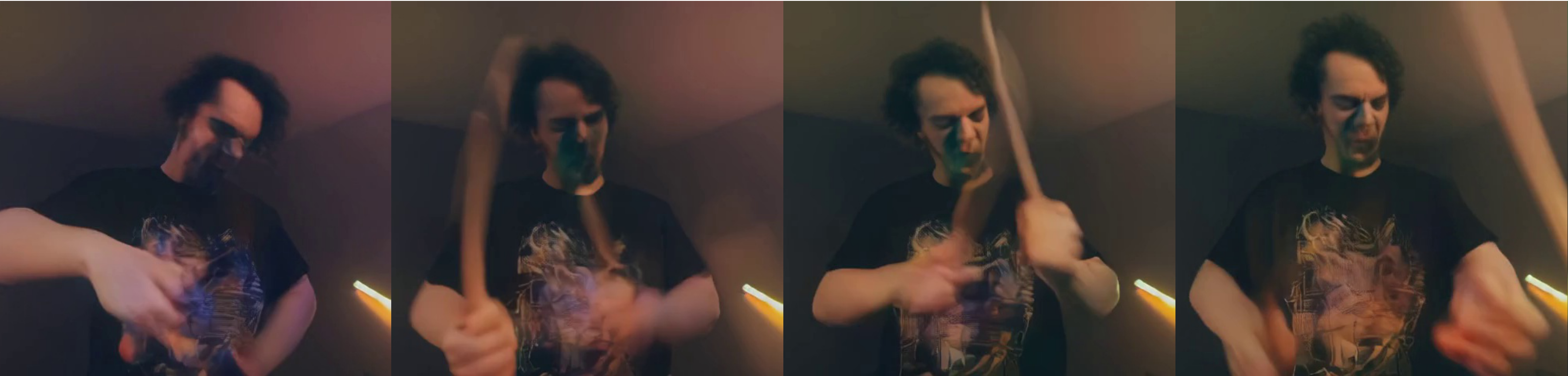} % replace with your image file
        \caption{Human: Good (Score=3) \cmark; MLLM: Good (Score=3) \cmark}
        \label{fig:air_drumming}
    \end{subfigure}

    \begin{subfigure}{\columnwidth}
        \centering
        \includegraphics[width=\columnwidth]{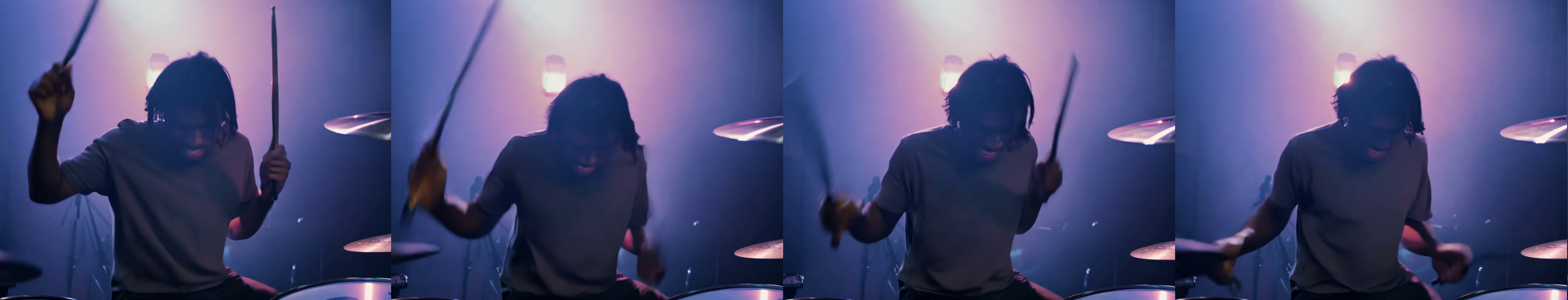} % replace with your image file
        \caption{Human: Good (Score=3) \xmark; MLLM: Moderate (Score=2) \cmark}
        \label{fig:actual drumming}
    \end{subfigure}
    \end{minipage}

    \caption{
\textbf{(a)} demonstrates perfect alignment with the prompt "Skis", where both MLLM and humans correctly assign high scores for accurate ski equipment representation.
\textbf{(b)} shows a critical discrepancy where humans incorrectly rate a snowboard as equivalent to skis, while MLLM appropriately penalizes this object mismatch, demonstrating superior object discrimination.
\textbf{(c)} illustrates the accurate generation of "air drumming" action, receiving rightfully high scores from both MLLM and human evaluators for proper action representation.
\textbf{(d)} reveals another human evaluation oversight where actual drum set playing is misjudged as equivalent to air drumming, while MLLM correctly identifies this semantic distinction and assigns a lower score.
}
    \label{fig:outperform_cases}
\end{figure}

% 就Confidence Interval Across Dimension那张表 分析一下
\begin{table*}[tbp]
\caption{
\textbf{Confidence interval across dimensions.} This table shows the mean difference between our evaluations and human evaluations after bootstrapping for $1000$ iterations over $100k$ pair of scores sampled with replacement. Positive score in mean difference indicates that HA-Video-Bench evaluations have higher sample mean as compared to human evaluations.
}

\scriptsize
\renewcommand\tabcolsep{2pt}
\renewcommand\arraystretch{1.2}
\renewcommand{\footnote}{\fnsymbol{footnote}} 
% \small
\setlength{\abovecaptionskip}{0.0cm}
\setlength{\belowcaptionskip}{-0.45cm}
\centering
\hspace*{0.00000000000000001cm}
% \scalebox{0.86}{%
\begin{tabular}{c|cccc|ccccc}

\hline

\hline

\hline

\hline

& \multicolumn{4}{c|}{\underline{\textit{\textbf{Video quality}}}} & \multicolumn{5}{c}{\underline{\textit{\textbf{Video-Condition Alignment}}}}  \\

Metrics&Imaging  & Aesthetic & Temporal & Motion & Video-text & Object-class & Color & Action & Scene   \\
& Quality & Quality & Consistency & Effects & Consist. & Consist. & Consist. & Consist. & Consist. \\
\hline

\hline
\text{Mean Difference} & $0.25$ & $0.18$ & $0.31$ & $-0.26$ & $0.19$ & $0.04$ & $0.05$ & $0.22$ & $0.11$ \\
99\% \text{ Confidence Interval} & $[0.23, 0.26]$ & $[0.16, 0.20]$ & $[0.28, 0.33]$ & $[-0.29, -0.23]$ & $[0.17, 0.21]$ & $[0.03, 0.05]$ & $[0.04, 0.06]$ & $[0.20, 0.23]$ & $[0.1, 0.12]$ \\
\hline

\hline

\hline

\hline

\end{tabular}
% }
% \vspace{0.25cm}

\label{tab:ci}
\end{table*}
% \input{table/prompts}
% In this section, we attach the prompt used to generate the evaluation results for the color dimension. 
% Tab~\ref{tab:description_prompt} shows the description prompt for step1.
% Tab~\ref{tab:questioning_prompts} are the questioning prompts for step2.
% Tab~\ref{tab:answering_prompt} shows the answering prompt for step3.
% Tab~\ref{tab:scoring_prompt} are the scoring prompt for step4.

\section{Potential Societal Impacts}
\label{sec:potential_societal_impacts} 
Leveraging this inherent cognitive capability of MLLM~\cite{wang2024comprehensive,wang2024gpt4video}, we can construct automated frameworks for evaluating video generation quality, fundamentally transforming the traditional paradigm that relies on human feedback~\cite{chen2024mllm,chen2024mj}. 
This breakthrough carries dual significance: firstly, it liberates human resources, freeing evaluators from the burden of manual annotation; secondly, through continuous and stable model feedback, it substantially accelerates the iterative optimization cycle of video generation technology.
From a long-term perspective, this technological advancement will significantly propel the development of virtual worlds~\cite{bruce2024genie}, establishing a more solid technical foundation for frontier applications such as the metaverse and digital humans. 
However, we must carefully acknowledge its potential societal impacts: as the authenticity of generated content continues to improve, how to effectively prevent and detect the spread of misinformation~\cite{shu2017fake}, and how to balance the social value of immersive content against the risks of excessive use~\cite{bojic2022culture}, are crucial issues that require joint attention and resolution from both academia and industry. 
This necessitates that while advancing technological innovation, we actively develop corresponding governance frameworks and ethical guidelines.

\section{Limitations and Future Work}
\label{sec:limitations_and_future_work}
While our approach demonstrates promising results, several limitations warrant acknowledgment. The current MLLM-based evaluation framework faces inherent constraints in perceiving dynamic elements and capturing fine-grained details. These limitations primarily manifest in the following aspects:
\begin{itemize}
    \item \textbf{Bounded capability}: The evaluation accuracy is fundamentally constrained by the inherent limitations of MLLMs in understanding complex temporal relationships and subtle visual nuances~\cite{zhang2024exploring,tang2023video}.
    \item \textbf{Model bias}: The evaluation framework may exhibit biases inherited from the pre-training data and architectural design of the underlying MLLMs.
\end{itemize}
To address these limitations, we identify several promising research directions:
\begin{itemize}
    \item \textbf{Differentiable metrics}: Development of differentiable evaluation metrics that can be directly integrated into the training pipeline, enabling end-to-end optimization of video generation models.
    
    \item \textbf{Optimization mapping}: Investigation of more robust methods for mapping evaluation results to concrete optimization strategies, potentially incorporating adaptive feedback mechanisms.
    
    \item \textbf{Enhanced temporal understanding}: Improving MLLMs' capability to capture and assess dynamic elements and temporal coherence in generated videos.
\end{itemize}
    
As MLLM technology continues to evolve, we anticipate significant improvements in evaluation accuracy and reliability. Future work should focus on developing more sophisticated architectures capable of capturing both global temporal dynamics and local visual details while maintaining computational efficiency.

\section{Annotation software}
Figure~\ref{fig:annotation_software} shows a web-based video annotation tool interface that supports video classification by groups and dimensions. The software provides clear navigation functions (Beginning, Previous, Next, End) to streamline the annotation workflow. The interface features a reminder panel on the left side that provides detailed evaluation guidelines for annotators.
\begin{figure}[h]
    \centering
    \includegraphics[width=1.0\linewidth]{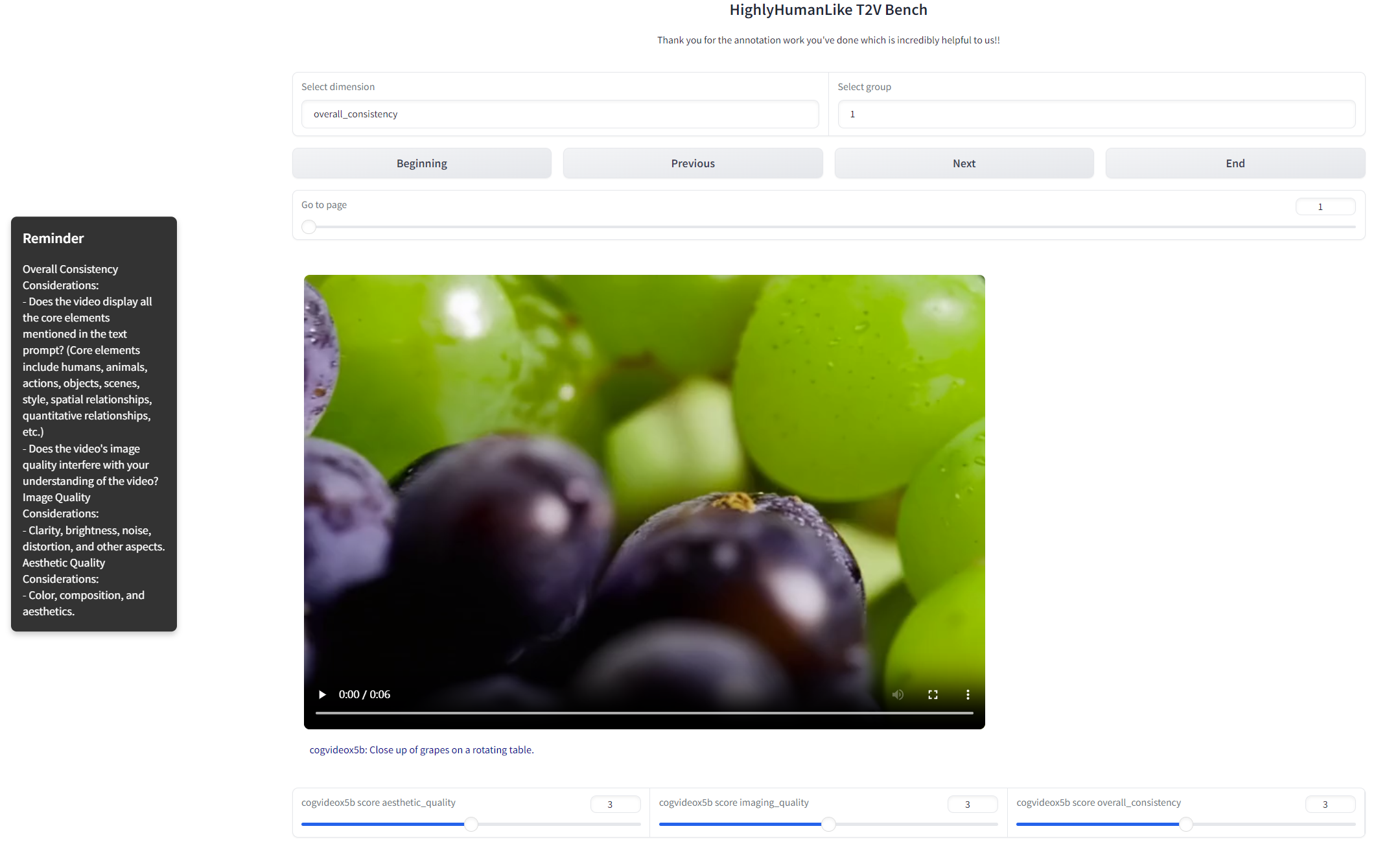}
    \caption{\textbf{Annotation software.}}
    \label{fig:annotation_software}
\end{figure}

\section{Examples of full prompt}
Taking the assessment of color consistency as an example, the process begins with GPT-4V analyzing video frames to generate initial descriptions, followed by two specialized assistants that ask a ``chain'' of questions on object-color accuracy and color relationships respectively. Then MLLM answers these questions.
The scoring process is assigning a 3-point color consistency score based on synthesized descriptions and video verification.

\clearpage
\onecolumn
\label{sec:example_of_full_prompt}
\paragraph{Video description}
To systematically evaluate color consistency, we design a specialized prompt for GPT-4V(ision) that instructs the model to analyze color-related attributes in videos. Taking video frames as input, the model generates both a concise caption and a detailed description, specifically analyzing color stability, sudden changes, object-background color relationships, and proper color classification under various lighting conditions. The following shows our carefully engineered system prompt:
\begin{minted}[fontsize=\footnotesize]{python}

<instructions>
### Task Description:
"""You are a video description expert, you need to focus on describing the colors in the video. You will receive an AI-generated video. 
Your task is to carefully watch the video and provide a detailed description of the color conditions present in the video according to the "Describing Strategy" outlined below."""

### Important Notes:
"""1.Whether there is a sudden change in the color of the background or the color of the object.
2.Is there a single color or multiple colors on the object.
3.Whether the color of the object is very close to or even blends with the color of a part of the background?
4.Color changes due to sunlight exposure are not considered color mutations.
5.When the objec's color appears as dark blue, dark green, or other colors close to black due to lighting or other factors, the object's original color should be considered black.
6.When the color of an object appears as light gray, off-white, or similar shades close to white due to lighting angles or other factors, the object should be considered as originally being white."""

### Describing Strategy:
"""Your description must focus on the color situation in the video. Please follow these steps:
1. Observe and identify the object in the video.
2. Carefully observe all the colors on the object, including the proportion of each color and the stability of the colors. Also describe the colors of the background and their stability. Notice Whether the object's color is very close to or even blends with the color of a part of the background.
3. Give a description of the entire video based on above observations.
4. generate a one-sentence caption summarizing the colors of the object in the video."""

### Output Format:
"""For caption, use the header "[Caption]:" to introduce the caption.
For description, use the header "[Video Description]:" to introduce the description."""

<example>
[Video Caption]:
(Here, describe the caption.)

[Video Description]:
(Here, describe the entire video.)
</example>

</instructions>

\end{minted}

\paragraph{Chain of query}
The evaluation assistant analyzes the consistency among text prompts, video descriptions, and concise captions, focusing on three dimensions: object recognition accuracy, color fidelity, and temporal color stability. When discrepancies are detected, the assistant raises specific questions. The following shows the prompt design for the evaluation assistant:
\begin{minted}[fontsize=\footnotesize]{python}

### Task Description:
"""You are an evaluation assistant whose role is to help the leader reflect on its descriptions of the generated video. 
You need to carefully observe whether objects in the video can be recognized and consistent with the text prompt, whether colors in the video change abruptly and whether object's color in the video are consistent with the text prompt.
Your task is to identify the differences in object accuracy, color accuracy and stability between the caption, video description, and text prompt.
You need to ask questions that highlight these differences. If these differences do not appear, do not ask questions."""

### Important Notes: 
"""1. Focus on whether the generated object is correct. If the video description doesn't indicate what the object is, you must ask.
2. Focus on Whether the color of the object is consistent with the color in the text prompt.
3. Focus on whether the color remains stable throughout the video (Color changes due to sunlight exposure are not considered color mutations.)
4. Your question must highlight specific differences where the color in the video does not match the text prompt, as shown in the example."""

### Questioning Strategy:
"""Based on the video caption and video description, compare it to the text prompt's object and color requirements. 
You're only allowed two questions at most. If there's no question, you can say I don't have a question.
Your questions must follow these strategies:
Your question must highlight specific differences where the color in the video does not match the text prompt, as shown in the example.

1. Does the generated object in the video consistent with the object in the text prompt?
2. Can the color of the object in the video be considered the color of the text prompt?
3. Whether there are sudden changes in colors in the video?

Example Questions:
"Whether the object in the video can be recognized?"
"Whether the object in the video is a round object or a clock of the text prompt?"
“Can the orange color of the cat in the video be considered the yellow color of the text prompt?”
“Whether there are sudden changes in colors in the video?"""

### Output Format:
"""You need to first analyze if there are any differences in object accuracy, color accuracy and stability between the caption, video description and the text prompt, and then decide whether to ask questions.
Your response should follow the format given in the example."""

<example>
[Your analysis]:
(Your analysis should be here)

[Your question]:
<question>
question:... 
I have no question.
</question>
</example>
\end{minted}
The second evaluation assistant examines two specific aspects of color relationships: the presence of additional colors beyond those specified in the text prompt, and the color contrast between objects and backgrounds. The following shows the prompt design for the second evaluation assistant:
\begin{minted}[fontsize=\footnotesize]{python}
### Task Description:
"""
You are an evaluation assistant whose role is to help the leader reflect on its descriptions of the generated video. 
Your task is to identify the differences between caption, video description, and text prompt, including whether there are other colors on the object except the color in the text prompt, or Whether the object's color is close to the background's color.
You need to ask questions that point out these differences. If these differences do not appear, do not ask questions."""

### Important Notes:
"""1. Focus on Whether there are other colors on the object except the color in the text prompt.
2. Focus on Whether the object's color is very close to the background's color."""

### Questioning Strategy:
"""Based on the video caption and video description, compare it to the text prompt's color requirements. 
You're only allowed two questions at most. If there's no question, you can say I don't have a question.
Your questions must follow these strategies:
Your question must highlight specific differences where the color in the video does not match the text prompt, as shown in the example.

1. Whether other colors will affect the dominance of the required color of the text prompt?
2. Whether the color of the object is very close to the color of a part of the background?

Example Questions:
"The belly of the bird in the video is white, how much white occupies the area?"
"at first glance, the required color is the main color?
“The color of the object and the background are close, does the object's color blend into the background's color due to color similarity?"""

### Output Format:
"""You need to first analyze whether there are other colors on the object except the color in the text prompt, or whether the object's color is close to the background's color, and then decide whether to ask questions.
Your response should follow the format given in the example."""

<example>
[Your analysis]:
(Your analysis should be here)

[Your question]:
<question>
question:... 
I have no question.
</question>

</example>
\end{minted}

\paragraph{Chain of answer}
The final evaluator synthesizes observations and addresses questions raised by the previous assistants. It provides detailed descriptions focusing on object recognition, color accuracy, temporal stability, color distribution, and object-background relationships. Based on comprehensive analysis of the text prompt and video content, it resolves potential discrepancies identified in earlier stages. The following shows the prompt design for the final evaluator:
\begin{minted}[fontsize=\footnotesize]{python}
### Task Description:
"""You are now a Video Evaluation Expert. Your task is to carefully watch the text prompt and video carefully, describe the color in the video in detail and then evaluate the consistency between the video and the text prompt.
Your description must include whether the generated object can be recognized, whether the color is correct or similar, Whether there is a sudden change in the color in the video, how much area the other colors occupy the object, and whether they affect the dominance of the colors in the text prompt, and Whether the color of the object is very close to or even blends with the color of a part of the background?
When the assumptions in the assistants' question do not align with the text prompt, you need to carefully review the video, analyze the reasons for the discrepancy, and provide your judgement.
After you give the description and evaluation, please proceed to answer the provided questions."""

### Important Notes:
"""1. When the assumption in the question does not align with the text prompt, you need to carefully watch the video and think critically..
2. Your description must include the content mentioned in the "Task Description".
3. Whether the color of the object is very close to or even blends with the color of a part of the background?
4. Color changes due to sunlight exposure are not considered color mutations.
5. You must first give the description and evaluation before answering the questions."""

### Output Format:
"""You need to provide a detailed description and evaluation, followed by answering the questions.
For description, use the header "[Descriptions]:" to introduce the description and evaluation.
For the answers, use the header "[Answers]:" to introduce the answers.

<example>
[Descriptions]:
(Here, provide a detailed description of the video and evaluation, focusing on the color conditions.)

[Answers]:
(Here, answer the questions.)
</example>"""

### Evaluation Steps:
"""Follow the following steps strictly while giving the response:
1. Carefully read the "Task Description" and "Important Notes".
2. Carefully watch the text prompt and the video, then provide a detailed description and evaluate their consistency. 
3. Answer the provided questions.
4. Display the results in the specified 'Output Format'."""
\end{minted}

\paragraph{Final scoring}
Taking two independent video descriptions as input, this evaluator first synthesizes an updated description and then assigns scores on a 3-point scale. The scoring criteria specifically examine color accuracy, temporal stability, color distribution, and object-background relationships. Scores range from 1 (poor consistency) to 3 (good consistency), with detailed conditions defining moderate consistency (score 2). The following shows the prompt design for the scoring system:
\begin{minted}[fontsize=\footnotesize]{python}
### Task Description:
"""You are now a Video Evaluation Expert responsible for evaluating the consistency between AI-generated video and the text prompt. 
You will receive two video informations. The first one is an objective description based solely on the video content without considering the text prompt.
The second description will incorporate the text prompt. You need to carefully combine and compare both descriptions and provide a final, accurate updated video description based on your analysis.
Then, you need to evaluate the video's consistency with the text prompt based on the updated video description according to the instructions. 

<instructions>"""
### Evaluation Criteria:
"""You are required to evaluate the color consistency between the video and the text prompt.
Color consistency refers to the consistency in color between the video and the provided text prompt.
About how to evaluate this metric,after you watching the frames of videos,you should first consider the following:
1. Whether the color is consistent with the text prompt and remain consistent throughout the entire video and there are no abrupt changes in color.
2. Whether the color is on the right object or background.
3. Whether the colors are similar but not exactly the same?"""

### Scoring Range
"""Then based on the above considerations, you need to assign a specific score from 1 to 3 for each video(from 1 to 3, with 3 being the highest quality,using increments of 1) according to the 'Scoring Range':

1. Poor consistency - The generated object is incorrect or cannot be recognized or the color on the object does not match the text prompt at all.(e.g., yellow instead of red).
2. Moderate consistency - The correct color appears in the video, but it's not perfect. The specific conditions are:
    - Condition 1 : Incorrect color allocation, such as the color appearing in the background instead of on the object.
    - Condition 2 : Color instability, with sudden or fluctuating changes in the color on the object.
    - Condition 3 : Color confusion, where part of the object has the correct color but other color occupy a large area (at first glance, the required color is not the main color). (e.g., a white vase is generated as a black and white striped vase.)
    - Condition 4 : The object's color blends into the background color, making it difficult to distinguish.
    - Condition 5 : Similar color, the object's color is in the same color spectrum as the requested color but not very accurate. (e.g., pink instead of purple, or yellow instead of orange.)
3.  Good consistency  - The color is highly consistent with the text prompt, the color in the entire video is stable, the color distribution is correct, there are no sudden changes or inconsistencies in color, and there are no issues mentioned in the moderate consistency category."""

###Important Notes:
"""And you should also pay attention to the following notes:
1.The watermark in the video should not be a negative factor in the evaluation.
2.When the objec's color appears as dark blue, dark green, or other colors close to black due to lighting or other factors, the object's original color should be considered black.
4.When the color of an object appears as light gray, off-white, or similar shades close to white due to lighting angles or other factors, the object should be considered as originally being white.
3.Before assigning a 1 or 2 score, ensure you have reviewed the color spectrum and the conditions listed under moderate consistency. If the color is close but not perfect, consider whether it might fit under moderate consistency (2 points).  """        

### Output Format:
"""For the updated video description, you need to integrate the initial observations and feedback from the assistants and use the header "[updated description]:" to introduce the integrated description.
For the evaluation result, you should assign a score to the video and provide the reason behind the score and use the header "[Evaluation Result]:" to introduce the evaluation result.

<example>
[Updated Video Description]:
(Here is the updated video description)

[Evaluation Result]:
([AI model's name]: [Your Score], because...)
</example>"""

### Evaluation Steps:
"""Follow the following steps strictly while giving the response:
1.Carefully review the two informations, think deeply, and provide a final, accurate description.
2.Carefully review the "Evaluation Criteria" and "Important Notes." Use these guidelines when making your evaluation.
3.Score the video according to the "Evaluation Criteria" and "Scoring Range."
4.Display the results in the specified "Output Format."
</instructions>"""
\end{minted}

\twocolumn
%%%%%%%%% REFERENCES


\end{document}